\documentclass[review, 10pt]{elsarticle}
\makeatletter
\def\ps@pprintTitle{%
	\let\@oddhead\@empty
	\let\@evenhead\@empty
	\def\@oddfoot{\centerline{\thepage}}%
	\let\@evenfoot\@oddfoot}
\makeatother

\usepackage[margin=2.5cm]{geometry}
\usepackage{setspace}
\usepackage{lmodern}
\usepackage{amsmath,amssymb}
\usepackage{amsfonts,amsthm}
\usepackage{lineno}
\usepackage{graphicx}
\usepackage{bm, bbm}
\usepackage{algorithm}
\usepackage{physics}
\usepackage{color}
\usepackage{pxfonts}
\usepackage[footnotesize,bf]{caption}
\usepackage{subcaption}
\usepackage{mathtools}
\usepackage{hhline}
\usepackage[T1]{fontenc}

\usepackage{placeins}

\graphicspath{{figures/}}
\usepackage{ulem}
\usepackage{tikz}
\usepackage{color}
\usepackage[]{xcolor}

\begin{document}
\begin{frontmatter}
\title{Monte Carlo PINNs: deep learning approach for forward and inverse problems involving high dimensional fractional partial differential equations}
\author[SNH]{Ling Guo}
\author[TJ]{Hao Wu}
\author[TJ]{Xiaochen Yu}
\author[LESC]{Tao Zhou}
\address[SNH]{Department of Mathematics, Shanghai Normal University, Shanghai, China}
\vspace{0.2cm}
\address[TJ]{School of Mathematical sciences, Tongji University, Shanghai, China}
\vspace{0.2cm}
\address[LESC]{Institute of Computational Mathematics and Scientific/Engineering
Computing, Academy of Mathematics and Systems Science, Chinese Academy
of Sciences, Beijing, China.}

\begin{abstract}
We introduce a sampling based machine learning approach, Monte Carlo physics informed neural networks (MC-PINNs), for solving forward and inverse fractional partial differential equations (FPDEs). As a generalization of physics informed neural networks (PINNs), our method relies on deep neural network surrogates in addition to a stochastic approximation strategy for computing the fractional derivatives of the DNN outputs. A key ingredient in our MC-PINNs is to construct an unbiased estimation of the physical soft constraints in the loss function. Our directly sampling approach can yield less overall computational cost compared to fPINNs proposed in \cite{pang2019fpinns} and thus provide an opportunity for solving high dimensional fractional PDEs. We validate the performance of MC-PINNs method via several examples that include high dimensional integral fractional Laplacian equations, parametric identification of time-space fractional PDEs, and fractional diffusion equation with random inputs. The results show that MC-PINNs is flexible and promising to tackle high-dimensional FPDEs.
\end{abstract}


\end{frontmatter}

\section{Introduction}\label{S:1}
Fractional partial differential equations (FPDEs) have been widely employed in modeling systems involving historical memory and long range interactions, such as solute transport in porous media \cite{DavidABenson2000}, viscoelastic constitutive laws \cite{Mainardi2010} and turbulent flow \cite{epps2018turbulence,Chen2006,song2016fractional}. In practice, it is in general impossible to obtain analytical solutions of complex FPDEs and thus many numerical methods have been developed. Among others, we mention, for example, finite difference methods, finite element methods and spectral methods. Readers are referred to \cite{AnnaLischke2018,li2019theory} and references therein for more details along this direction. The main challenge for numerically solving FPDEs is routed in the expensive computational cost and high memory requirements due to the nonlocal property and singularity of the fractional derivatives. Especially for inverse problems modeled by FPDEs, one needs to identify the fractional derivative order or other parameters from the observation data via expensive forward FPDEs solvers.

Recently, machine learning techniques has been widely adopted to solve forward and inverse partial differential equations \cite{ERev}. Among these are Gaussian process regression ~\cite{graepel2003, sarkka2011, bilionis2016, Raissi_nonlinear, GuofeiPang2018, xiuyang_GP} and deep neural networks (DNNs)~\cite{Lagaris1997, Lagaris2000, Yinglexing17, MaziarParisGK17_1,EYu2018,ZangBaoYeZhou2020,LiaoMing2021,HWZ2022}. In this work, we focus on physics-informed neural networks (PINNs) that was first introduced in ~\cite{MaziarParisGK17_1,Raissi2019}. The key idea of PINNs is to include physics law (i.e., the PDE) into a deep neural network (DNN) that shares parameters with the DNNs-surrogate for the solution of the PDE. This strategy enables us to use less data during the training process and can better express the physical law. We can thus predict the system state unlike deep learning approach driven solely by data. PINNs is simple to implement and easy for coding, and has been shown to be successful for diverse forward and inverse problems in physics and fluid mechanics~\cite{M.Raissi2019, raissi2020hidden}.

The success of DNNs-based approaches for PDEs (such as PINNs) relies on well developed tools such as automatic differentiation for dealing with integer-order partial differential equations. However, this is not true for PDEs with nonlocal operators (such as fractional PDEs). To this end, in \cite{pang2019fpinns}, the authors extended PINNs to fractional PINNs (fPINNs) for solving space-time fractional advection-diffusion equations. The main idea for fPINNs is to use automatic differentiation for the integer-order operators, while numerical discretization scheme such as finite difference is employed for the fractional derivative of the neural network output. For example, the directional fractional Laplacian of the neural network output can be computed by combining the shifted vector Gr\"unwald--Letnikov (GL)
formula and quadrature rules. These combination leads to exhaustively cost as the physical dimension increasing and finally makes fPINNs infeasible for solving high dimensional fractional PDEs. We also mention that a nonlocal-PINNs for a parameterized nonlocal universal Laplacian operator is investigated in \cite{pang2020npinns}.

In this paper, we shall propose a Monte Carlo sampling based PINN, named MC-PINN, for solving forward and inverse fractional partial differential equations. The main idea of our MC-PINNs lies in that 1) we compute the fractional derivative of the DNN-output via a Monte Carlo. 2) During the training step, an unbiased estimate of the physics based loss function is designed to obtain the optimal DNNs-parameters. Compared to fPINNs, Our approach admits the following main advantages:
\begin{itemize}
\item Unlike fPINNs, the fractional derivative of the DNNs-output is computed via a directly sampling approach instead of using traditional schemes such as the finite difference method, which alleviate the computational cost and is promising for high dimensional problems.
\item The MC-PINNs model can also be used for solving parametric FPDEs where the inputs parametric is random and leads to uncertainty quantification problems.
\end{itemize}

We demonstrate the effectiveness of MC-PINNs by solving forward and inverse high dimensional space-time fractional PDEs. Classical methods have been developed for 3D space-fractional ADEs, but most of them focus on the Riesz space fractional derivative \cite{Wang2014,MengZhao2018}, which differs from the hyper-singular integral fractional Laplacian we considered here. While there are some works focusing on 1D/2D inverse space-time fractional PDEs with fractional Laplacian \cite{Miller2013,Minden2018}, seldom research has been conducted for 3D problem. Here, we consider high-dimensional inverse fractional Laplacian problem defined on a bounded domain which is the bottleneck for classical numerical methods.

The organization of this paper is as follows. In Section \ref{S:2}, we set up the forward and inverse FPDEs. In Section \ref{S:3}, we introduce the a directly sampling approach to compute the fractional derivative of the DNNs-output, and this is followed by our main algorithm -- the MC-PINNs for solving fractional PDEs. In Section \ref{S:4}, we first present a detailed study of the accuracy and performance of our MC-PINNs model for space fractional Laplacian operator in a bounded domain, including ten dimensional problems. Then we present the simulation results for parameters identification in time-space fractional partial differential equations. Finally, we show the flexibility of MC-PINNs  for solving FPDEs with random inputs. We finally conclude the paper in Section \ref{S:5}.

\section{Notations and problem setup}\label{S:2}
On a bounded spatial domain $\Omega \subset \mathbb{R}^d$, we consider the following high-dimensional fractional advection-diffusion equation
\begin{equation}\label{eqn:SDE}
\begin{array}{rcl}
\mathcal{L}[u(x,t)]:=\frac{\partial^{\gamma}u(x,t)}{\partial t^{\gamma}}+c(-\triangle)^{\alpha/2}u(x,t)+v\cdot\nabla u(x,t) & = & f(x,t),\quad (x,t)\in \Omega\times (0,T], \\
u(x,0) & = & g(x), \quad\quad  x\in \Omega,\\
u(x,t) & = & 0, \quad\qquad (x,t)\in (\mathbb{R}^d \backslash \Omega)\times (0,T]   ,
\end{array}
\end{equation}
where $c$ is the diffusion coefficient (deterministic or random variable), $v$ is the mean-flow velocity, and we assume zero boundary condition for simplicity. Here $\frac{\partial^{\gamma}}{\partial t^{\gamma}}$ is the Caputo-type time-fractional derivatives of order $\gamma$ defined by:
\begin{equation}\label{eqn:FPDE}
\begin{gathered}
    \frac{\partial^{\gamma}u(x,t)}{\partial t^{\gamma}}\triangleq\frac{1}{\Gamma(1-\alpha)}\int_0^t (t-\tau)^{-\gamma}\frac{\partial u(x,\tau)}{\partial \tau}d\tau, \quad 0< \gamma< 1.
\end{gathered}
\end{equation}
where $\Gamma(\cdot)$ is the Gamma function. The fractional Laplacian operator we considered in this paper is defined via the hyper-singular integral, i.e. \cite{AnnaLischke2018}:
\begin{equation}\label{eqn:Flaplacian}
   (-\Delta)^{\alpha/2}u(x) \triangleq C_{d,\alpha}\text{P.V.}\int_{\mathbb{R}^d}\frac{u(x)-u(y)}{\left\Vert x-y\right\Vert _{2}^{d+\alpha}}\mathrm{d}y , \quad 0<\alpha<2,
\end{equation}
where \text{P.V.} denotes the principle value of the integral and $C_{d,\alpha}$ is given by
\begin{equation}\label{eqn:Flaplaciancosntant}
   C_{d,\alpha}= \frac{2^{\alpha}\Gamma(\frac{\alpha+d}{2})}{\pi^{d/2}|\Gamma(-\alpha/2)|}.
\end{equation}
We consider two types of FPDE problems in this work:
\begin{itemize}
    \item \textit{Forward problem}: We know exactly the fractional order $\alpha$ and $\gamma$, the diffusion coefficient $c$, the velocity $v$ and the force term $f$, as well as the boundary/initial state to provide boundary/intial conditions of $u$, and our quantity of interest (QoI) is $u(x,t)$;
    \item \textit{Inverse problem}: In addition to the boundary/intial time, we have a limited number of extra $u$-sensors that can be placed in the time-space domain $\Omega\times (0,T]$ to collect data, while we are interested in inferring $\alpha$, $\gamma$, $c$, $v$ and the entire information of the solution $u(x,t)$.
\end{itemize}

Our main goal of this paper is to address both types of problems via a deep learning approach. To this end, we shall construct a neural network surrogates $u_{NN}(x,t)$ for the solution $u(x,t)$ of (\ref{eqn:SDE}) and then optimize the DNNs-parameters such that the approximation $u_{NN}(x,t)$ satisfy both the observed data on $u$-sensors and $f$-sensors. Our main innovation is the formulation of an unbiased estimation of the mean square equation loss function for the FPDEs, which results in dramatically reduced computational complexity (compared to fPINNs in \cite{pang2019fpinns}) and can be used to solve high-dimensional FPDEs. 
\section{Methodology}
\label{S:3}
\subsection{Physics-Informed Neural Network (PINNs)}
\label{S:3-1}
In this part, we first briefly review the main idea of DNNs-based approach for solving integer-order partial differential equations~\cite{Lagaris1997, Lagaris2000, Raissi2019}. To this end, we consider the following problem:
\begin{equation}\label{eqn:PDE_PINNs}
\begin{array}{rcl}
\mathcal{L}_{\lambda}[u(x,t)]:=\frac{\partial u(x,t)}{\partial t} + \frac{\partial^2 u(x,t)}{\partial x^2} & = & f(x,t),\quad (x,t)\in \Omega\times (0,T], \\
u(x,0) & = & g(x), \qquad x\in \Omega,\\
u(x,t) & = & 0, \quad\qquad x\in \partial\Omega,
\end{array}
\end{equation}
where $u(x,t)$ is the solution and $\lambda$ denotes the problem parameters.

\begin{figure}[htbp]
	\centering
	\includegraphics[width=0.65\linewidth]{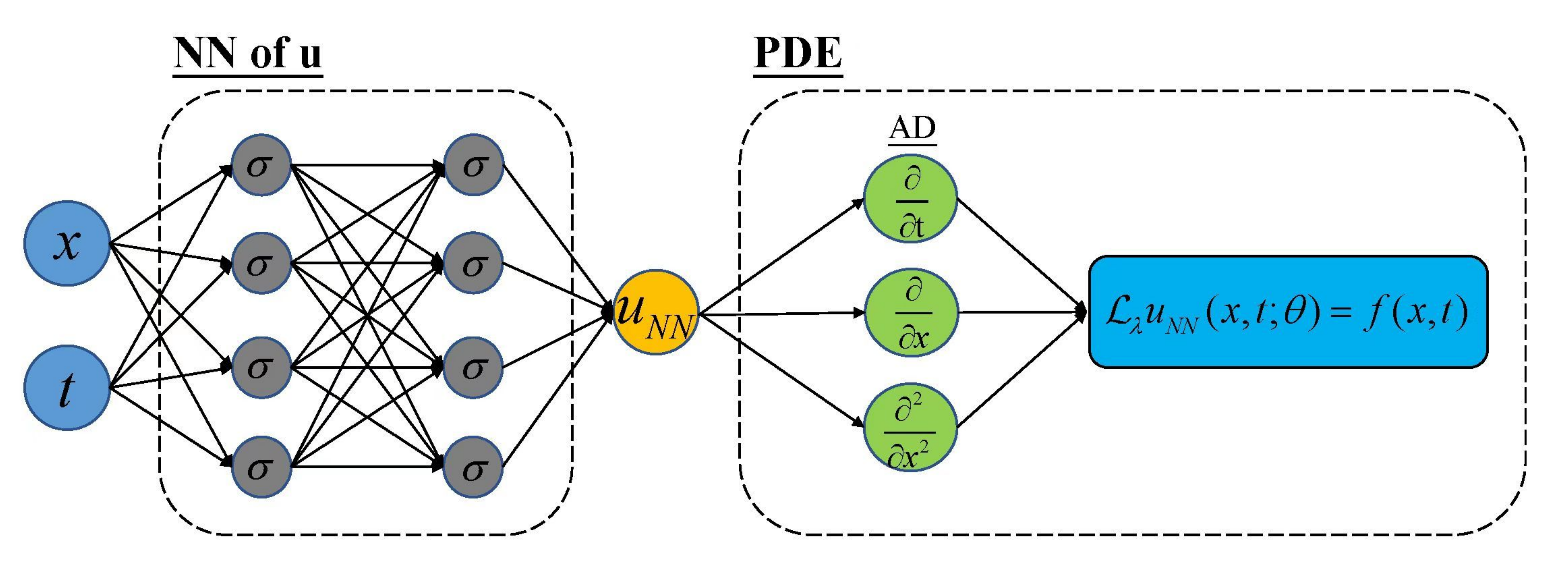}
	\caption{Schematic of the PINNs for solving partial differential equations.}
	\label{fig:nn_sketch}
\end{figure}

The PINNs approach solve the above forward PDE problems via constructing a DNN surrogate $u_{NN}(x,t;\theta)$, parametrized by $\theta$, of the solution $u(x,t)$. More precisely, $u_{NN}$ takes the coordinate $x,t$ as the input and outputs a vector that has the same dimension as $u(x,t)$. This surrogate $u_{NN}$ is then substituted into Eq.~(\ref{eqn:PDE_PINNs}) via automatic differentiation, which is conveniently integrated
in many machine learning packages to obtain $$f_{NN}=\mathcal{L}_{\lambda}[u_{NN}].$$  Assume that we have the training data set $\mathcal{D}=(\mathcal{D}_f, \mathcal{D}_g, \mathcal{D}_u)$, where $$\mathcal{D}_f=\{x_i,t_i,f_i\}_{i=1}^{N_f},\quad  \mathcal{D}_g=\{x_i,t_i,g_i\}_{i=1}^{N_g},\quad \mathcal{D}_u=\{x_i,t_i,u_i\}_{i=1}^{N_u},$$ and  $$f_i=f(x_i,t_i), \quad g_i=g(x_i,t_i), \quad u_i=u(x_i,t_i).$$
Notice that the data locations in the physical domain of $f$ ,$g$ and $u$ are usually different in general.
We use the same symbol $(x_i,t_i)$ for simplicity here since there is no misunderstanding.
Then at the training stage, the DNN-parameters $\theta$ are optimized, denoted by $\hat{\theta}$, by fitting the data set $\mathcal{D}=(\mathcal{D}_f, \mathcal{D}_g, \mathcal{D}_u)$ via minimizing the following loss function:
\begin{equation}\label{eqn:PINNs_loss}
  \mathcal{LOSS}(\theta)=\frac{1}{N_u}\sum_{i=1}^{N_u}\left[u_{NN}(x_i,t_i;\theta)-u_i\right]^2+\frac{1}{N_f}\sum_{i=1}^{N_f}
  \left[f_{NN}(x_i,t_i;\theta)-f_i\right]^2+\frac{1}{N_g}\sum_{i=1}^{N_g}\left[u_{NN}(x_i,0;\theta)-g_i\right]^2.
\end{equation}
Upon determination of $\hat{\theta}$, $u_{NN}(x,t;\hat{\theta})$ can be evaluated at any $(x,t)\in \Omega$. Fig.\ref{fig:nn_sketch} shows a sketch of the PINNs.

Inspired by the above PINNs approach, an extended version, named fractional PINNs (fPINNs), is established in \cite{pang2019fpinns} for solving fractional partial differential equations. However, automatic differentiation can not be used directly for fractional derivatives. fPINNs thus employ the traditional discrete techniques, such as the finite difference method, for the fractional differential operators to obtain the loss function of PINNs. Obviously, this approach needs many auxiliary points for each training points and thus suffers from the curse of dimensionality for high-dimensional problems. To tackle this problem, we shall employ a stochastic approximation strategy to compute the fractional derivatives of the DNNs-output and establish our MC-PINNs strategy for solving forward and inverse problems of fractional PDEs.

\subsection{Monte Carlo Physics-Informed Neural Networks (MC-PINNs)}
\begin{figure}[htbp]
	\centering
	\includegraphics[width=0.65\linewidth]{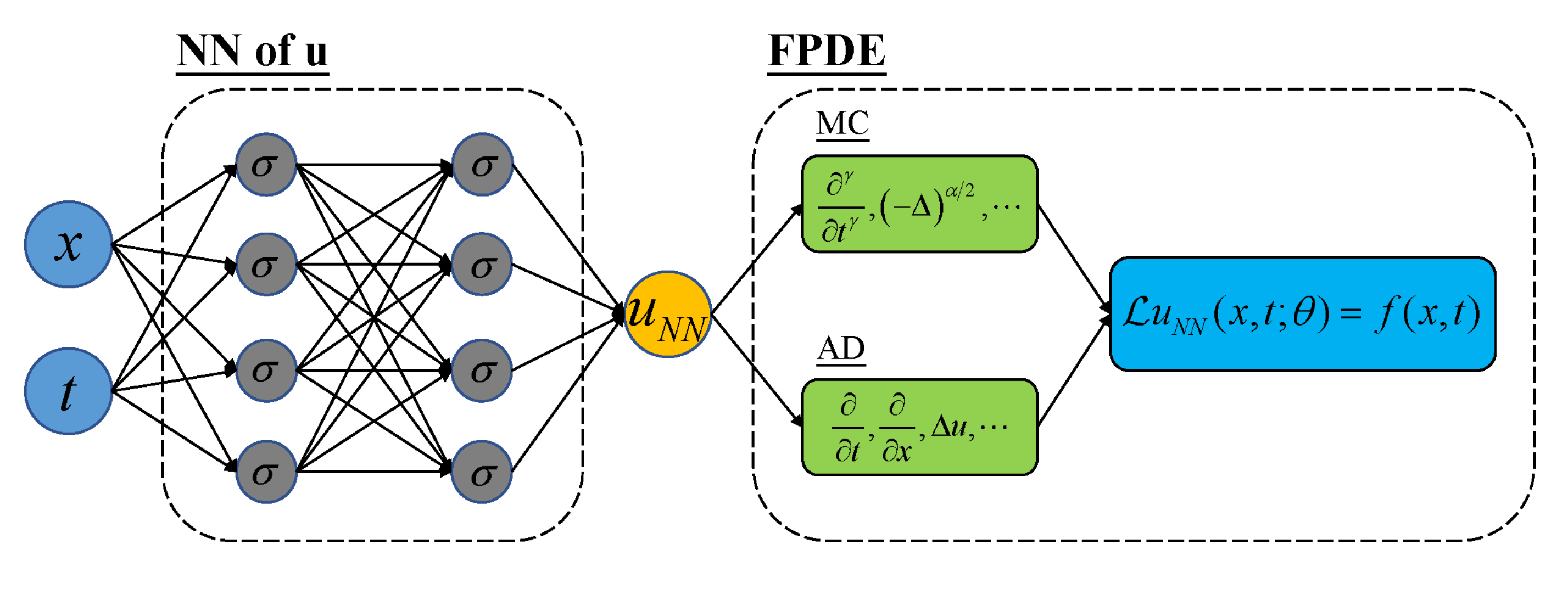}
	\caption{Schematic of the MC-PINNs for solving forward and inverse fractional partial differential equations.}
	\label{fig:MC-pinn_sketch}
\end{figure}

In this section we formalize the algorithm of solving  Eq.~(\ref{eqn:SDE}).
Given the training data set $\mathcal{D}=(\mathcal{D}_f, \mathcal{D}_g, \mathcal{D}_u)$, we define the loss function as
\begin{equation}\label{eqn:loss}
\mathcal{LOSS}(\theta)=w_{equ}L_{equ}(\theta)+w_{g}L_{g}(\theta)+w_{u}L_{u}(\theta),
\end{equation}
where
\begin{equation}\label{eqn:sdeloss1}
L_{equ}(\theta) =  \| L[u_{NN}(x,t;\theta)]-f(x,t)\|^2, \quad L_{g}(\theta) = \frac{1}{N_g}\sum_{i=1}^{N_g}[u_{NN}(x_i,0;\theta)-g_i]^2, \quad
L_{u}(\theta)=\frac{1}{N_u}\sum_{i=1}^{N_u}[u_{NN}(x_i,t_i;\theta)-u_i]^2.
\end{equation}
Here $w_{equ}$, $w_{g}$ and $w_{u}$ are the weights of the different parts of the loss function. The schematic of the MC-PINNs method is shown in Fig.~\ref{fig:MC-pinn_sketch}. In the next section, we shall give details for constructing the equation loss $L_{equ}$ and minimization of the loss function $\mathcal{LOSS}$. We first consider a directly sampling Monte Carlo method for approximating the nonlocal operators in $\mathcal{L}$ that can not be automatically differentiated, for example, $\frac{\partial^{\gamma}}{\partial t^{\gamma}}$ and $(-\Delta)^{\alpha/2}$ for $\gamma\in (0,1)$, $\alpha\in (0,2)$.

\subsubsection{Stochastic approximation of fractional operators}
To compute the fractional Laplacian of the DNN-output $u_{NN}(x,t;\theta)$ with $\alpha\in (0,2)$, we first divide the integral into 
integrals over a neighborhood
$B_{r_0}(x)=\{y|\Vert y-x\Vert\le r_0\}$ around $x$ and its complement as the following:
\begin{equation}\label{eqn:Flaplacian1}
 (-\Delta)^{\alpha/2}u_{NN}(x) = C_{d,\alpha}\bigg (\int_{y\in B_{r_{0}}(x)}\frac{u_{NN}(x)-u_{NN}(y)}{\left\Vert x-y\right\Vert _{2}^{d+\alpha}}\mathrm{d}y + \int_{y\notin B_{r_{0}}(x)}\frac{u_{NN}(x)-u_{NN}(y)}{\left\Vert x-y\right\Vert _{2}^{d+\alpha}}\mathrm{d}y\bigg ).
\end{equation}
Here we omit $t$ and $\theta$ and denote $u_{NN}(x,t;\theta)$ as $u_{NN}(x)$ for notation simplicity.

We can re-write the first part as
\begin{equation}\label{eqn:Flaplacian11}
\begin{aligned}
\int_{y\in B_{r_{0}}(x)}\frac{u_{NN}(x)-u_{NN}(y)}{\left\Vert x-y\right\Vert _{2}^{d+\alpha}}\mathrm{d}y & = \int_{\left\Vert y\right\Vert _{2}\le r_{0}}\frac{u_{NN}(x)-u_{NN}(x-y)}{\left\Vert y\right\Vert _{2}^{d+\alpha}}\mathrm{d}y\\
 & =  \frac{1}{2}\int_{\left\Vert y\right\Vert _{2}\le r_{0}}\frac{2u_{NN}(x)-u_{NN}(x-y)-u_{NN}(x+y)}{\left\Vert y\right\Vert _{2}^{d+\alpha}}\mathrm{d}y\\
 & =  \frac{1}{2}\int_{S^{d-1}}\int_{0}^{r_{0}}\frac{2u_{NN}(x)-u_{NN}(x-r\xi)-u_{NN}(x+r\xi)}{r^{1+\alpha}}\mathrm{d}r\mathrm{d}\xi\\
 & =  \frac{1}{2}\int_{S^{d-1}}\int_{0}^{r_{0}}\frac{2u_{NN}(x)-u_{NN}(x-r\xi)-u_{NN}(x+r\xi)}{r^{2}}r^{1-\alpha}\mathrm{d}r\mathrm{d}\xi\\
 & =  \frac{\left|S^{d-1}\right|r_{0}^{2-\alpha}}{2\left(2-\alpha\right)}\mathbb{E}_{\xi,r\sim f_{I}(r)}\left[\frac{2u_{NN}(x)-u_{NN}(x-r\xi)-u_{NN}(x+r\xi)}{r^{2}}\right],
\end{aligned}
\end{equation}
where $\xi$ is uniformly distributed on the the unit $(d-1)$-sphere $S^{d-1}$,
$|S^{d-1}|$ denotes the surface area of $S^{d-1}$,
\[
f_{I}(r)=\frac{2-\alpha}{r_{0}^{2-\alpha}}r^{1-\alpha}\cdot1_{r\in[0,r_{0}]},
\]
and $r$ can be sampled as
\begin{equation}\label{eqn:rI_sample}
r/r_0 \sim \mathrm{Beta}(2-\alpha,1).
\end{equation}

Notice that we have 
\[
\lim_{r\to0}\frac{2u_{NN}(x)-u_{NN}(x-r\xi)-u_{NN}(x+r\xi)}{r^{2}}=\left.\frac{\partial^{2}u_{NN}(x+r\xi)}{\partial r^{2}}\right|_{r=0},
\]
and this finite difference type approximation may suffer from the rounding
error and yield numerical instability for an extremely small $r$.
Thus we utilize the following approximation in practice in (\ref{eqn:Flaplacian1})
\[
\int_{y\in B_{r_{0}}(x)}\frac{u_{NN}(x)-u_{NN}(y)}{\left\Vert x-y\right\Vert _{2}^{d+\alpha}}\mathrm{d}y\approx\frac{\left|S^{d-1}\right|r_{0}^{2-\alpha}}{2\left(2-\alpha\right)}\mathbb{E}_{\xi,r\sim f_{I}(r)}\left[\frac{2u_{NN}(x)-u_{NN}(x-r_{\epsilon}\xi)-u_{NN}(x+r_{\epsilon}\xi)}{r_{\epsilon}^{2}}\right],
\]
with $r_{\epsilon}=\max\{\epsilon,r\}$, where $\epsilon>0$ is a small positive number.

Similarly, for the second part in (\ref{eqn:Flaplacian1}) we have
\begin{equation}\label{eqn:Flaplacian12}
\begin{aligned}
\int_{y\notin B_{r_{0}}(x)}\frac{u_{NN}(x)-u_{NN}(y)}{\left\Vert x-y\right\Vert _{2}^{d+\alpha}}\mathrm{d}y & =  \int_{\left\Vert y\right\Vert _{2}\ge r_{0}}\frac{u_{NN}(x)-u_{NN}(x-y)}{\left\Vert y\right\Vert _{2}^{d+\alpha}}\mathrm{d}y\\
 & =  \int_{S^{d-1}}\int_{r_{0}}^{\infty}\frac{u_{NN}(x)-u_{NN}(x-r\xi)}{r^{1+\alpha}}\mathrm{d}r\mathrm{d}\xi\\
 & =  \frac{\left|S^{d-1}\right|r_{0}^{-\alpha}}{2\alpha}\mathbb{E}_{\xi,r\sim f_{O}(r)}\left[2u_{NN}(x)-u_{NN}(x-r\xi)-u_{NN}(x+r\xi)\right],
\end{aligned}
\end{equation}
where
\[
f_{O}(r)=\alpha r_{0}^{\alpha}r^{-1-\alpha}1_{r\in[r_{0},\infty)},
\]
and $r$ can be sampled via
\begin{equation}\label{eqn:ro_sample}
r_0/r \sim \mathrm{Beta}(\alpha,1).
\end{equation}

Consequently, by combining (\ref{eqn:Flaplacian11}) and (\ref{eqn:Flaplacian12}), the fractional Laplacian of the surrogate $u_{NN}$ can be calculated via the following
approximation:
\begin{equation}\label{eqn:Flaplacianmc}
\begin{aligned}
\left(-\Delta\right)^{\alpha/2}u_{NN}(x) & =  C_{d,\alpha}\frac{\left|S^{d-1}\right|r_{0}^{2-\alpha}}{2\left(2-\alpha\right)}\mathbb{E}_{\xi,r_I\sim f_{I}(r)}\left[\frac{2u_{NN}(x)-u_{NN}(x-r_{\epsilon}\xi)-u_{NN}(x+r_{\epsilon}\xi)}{r_{\epsilon}^{2}}\right]\\
 &   +C_{d,\alpha}\frac{\left|S^{d-1}\right|r_{0}^{-\alpha}}{2\alpha}\mathbb{E}_{\xi,r_I\sim f_{O}(r)}\big[2u_{NN}(x)-u_{NN}(x-r_o\xi)-u_{NN}(x+r_o\xi)\big].
\end{aligned}
\end{equation}
Here $r_{\epsilon}=\max\{\epsilon,r_I\}$, and $r_I$ is sampled according to Eq.~(\ref{eqn:rI_sample}) while $r_o$ is sampled via Eq.~(\ref{eqn:ro_sample}).

To approximate the time fractional derivative of the DNN-ouput $u_{NN}(x,t;\theta)$ in equation (\ref{eqn:SDE}) for $\gamma\in (0,1)$, we adopt again the stochastic approximation via MC sampling as follows:
\begin{equation}\label{eqn:caputo}
\begin{aligned}
\frac{\partial^{\gamma}u_{NN}(x,t)}{\partial t^{\gamma}}& =\int_{0}^{t}(t-\tau)^{-\gamma}\frac{\partial}{\partial\tau}{u}_{NN}(x,\tau)\mathrm{d}\tau \\
& =  \gamma\int_{0}^{t}\tau^{-\gamma}\frac{u_{NN}(x,t)-u_{NN}(x,t-\tau)}{\tau}\mathrm{d}\tau+\frac{u_{NN}(x,t)-u_{NN}(x,0)}{t^{\gamma}}\\
 & =  \frac{\gamma}{\left(1-\gamma\right)}t^{1-\gamma}\mathbb{E}_{\tau\sim f_{I,t}}\left[\frac{u_{NN}(x,t)-u_{NN}(x,t-\tau)}{\tau}\right]+\frac{u_{NN}(x,t)-u_{NN}(x,0)}{t^{\gamma}}\\
 & \approx  \frac{\gamma}{\left(1-\gamma\right)}t^{1-\gamma}\mathbb{E}_{\tau\sim f_{I,t}}\left[\frac{u_{NN}(x,t)-u_{NN}(x,t-\tau_{\epsilon}t)}{\tau_{\epsilon}t}\right]+\frac{u_{NN}(x,t)-u_{NN}(x,0)}{t^{\gamma}},
\end{aligned}
\end{equation}
where $f_{I,t}(\tau)={\left(1-\gamma\right)}\tau^{-\gamma}\cdot1_{\tau\in[0,1]}$, and $\tau$ can be sampled via
\begin{equation}\label{eqn:tau_sample}
\tau \sim \mathrm{Beta}(1-\gamma,1).
\end{equation}
Moreover, $\tau_{\epsilon}=\max\{\tau,\epsilon_{t}t^{-1}\}$,
and $\epsilon_{t}$ is a small positive number which we will specify it in the numerical examples.

Based on the above stochastic approximations for the space and time fractional derivative (\ref{eqn:Flaplacianmc}) and  (\ref{eqn:caputo}), along with the automatic differentiation for the integer-order derivative, we can finally obtain the approximation for
$ L[u_{NN}(x,t;\theta)]$ as follows
\begin{equation}\label{eqn:ADE_NN}
\begin{aligned}L[u_{NN}(x,t;\theta)] & \approx\widehat{L}[u_{NN}(x,t;\theta);\epsilon,\epsilon_{t},\tau,\xi,r_{I},r_{o}]\\
 & =\frac{\gamma}{\left(1-\gamma\right)}t^{1-\gamma}\cdot\frac{u_{NN}(x,t;\theta)-u_{NN}(x,t-\tau_{\epsilon}t;\theta)}{\tau_{\epsilon}t}+\frac{u_{NN}(x,t;\theta)-u_{NN}(x,0;\theta)}{t^{\gamma}}\\
 & +C_{d,\alpha}\frac{\left|S^{d-1}\right|r_{0}^{2-\alpha}}{2\left(2-\alpha\right)}\cdot\frac{2u_{NN}(x,t;\theta)-u_{NN}(x-r_{\epsilon}\xi,t;\theta)-u_{NN}(x+r_{\epsilon}\xi,t;\theta)}{r_{\epsilon}^{2}}\\
 & +C_{d,\alpha}\frac{\left|S^{d-1}\right|r_{0}^{-\alpha}}{2\alpha}\cdot\left(2u_{NN}(x,t;\theta)-u_{NN}(x-r_{o}\xi,t;\theta)-u_{NN}(x+r_{o}\xi,t;\theta)\right)\\
 & +v\cdot\nabla u_{NN}(x,t;\theta),
\end{aligned}
\end{equation}
where $r_I,r_o,\tau$ are distributed according to $f_I,f_O,f_{I,t}$, $r_\epsilon=\max\{r_I,\epsilon\}$, and $\xi$ is drawn from the uniform distribution on the sphere $S^{d-1}$.

\subsubsection{Unbiased estimation of the equation loss}
We now describe how to evaluate the unbiased estimates of the equation loss $L_{equ}$ in Eq.~(\ref{eqn:loss}). Based on the stochastic approximation for the fractional PDE operator Eq.~(\ref{eqn:ADE_NN}), we can obtain the unbiased estimates of $L_{equ}$ by implementing the following steps:
\begin{enumerate}
    \item Sample $x_i$ and $t_i$ uniformly from $\Omega\subset\mathbb{R}^d$ and the time domain respectively, $i=1,\cdot\cdot\cdot,N_u$.
    \item Given small numbers $\epsilon$ and $\epsilon_{t}$, for each residual point $(x_i,t_i)$, we sample two groups random parameters $\tau$, $\xi$, $r_I$, $r_o$ according to their prior distributions and denote them by $\{\tau_{i}, \xi_{i}, {r_I}_{i},{r_o}_{i}\}_{i=1}^{m}$, $\{\tau^{\prime}_{i}, \xi^{\prime}_{i}, {r_I}^{\prime}_{i},{r_o}^{\prime}_{i}\}_{i=1}^{m}$ respectively. Here $m$ represents the number of samples.

   \item Then we calculate
\begin{equation}\label{eqn:empericalloss}
\hat{L}_{equ}(\theta)=\frac{1}{mN_u}\sum_{i,j}\widehat{L}\big[u_{NN}(x_{i},t_{i};\theta);\epsilon,\epsilon_{t},\tau_{j},\xi_{j},r_{Ij},r_{oj}\big]
\cdot \widehat{L}\big[u_{NN}(x_{i},t_{i};\theta);\epsilon,\epsilon_{t},\tau_{j}^{\prime},\xi_{j}^{\prime},r_{Ij}^{\prime},r_{oj}^{\prime}\big]
\end{equation}
according to Eq.~(\ref{eqn:ADE_NN}).
\end{enumerate}
It can be seen from the above analysis that $\mathbb E[\hat{L}_{equ}(\theta)]=L_{equ}(\theta)$ if $\epsilon=\epsilon_t=0$ and the rounding error can be ignored.
Now we are ready to put Eq.~(\ref{eqn:empericalloss}) into Eq.~(\ref{eqn:loss}) to formulate the total loss function. We summarize the MC-PINNs method in Algorithm 1.

\begin{algorithm}[H]
\begin{itemize}
\item\textbf{1.} Specify the training data set
\begin{equation*}
\mathcal{D}=(\mathcal{D}_f, \mathcal{D}_g, \mathcal{D}_u), \text{ where } \mathcal{D}_f=\{x_i,t_i,f_i\}_{i=1}^{N_f},  \mathcal{D}_g=\{x_i,t_i,g_i\}_{i=1}^{N_g}, \mathcal{D}_u=\{x_i,t_i,u_i\}_{i=1}^{N_u}
\end{equation*}
\item\textbf{2.} Sample $N$ snapshots from the above training data
\item\textbf{3.} Calculate the loss $\mathcal{LOSS}(\theta)=w_{equ}\hat L_{equ}(\theta)+w_{g}L_{g}(\theta)+w_{u}L_{u}(\theta)$ via Eqs.~(\ref{eqn:loss}) and (\ref{eqn:empericalloss})
\item\textbf{4.} Let $\theta\leftarrow \text{Adam}(\theta-\eta\frac{\partial \mathcal{LOSS}(\theta)}{\partial \theta})$ to update all the involved parameters $W$ in (\ref{eqn:loss}), $\eta$ is the learning rate
\item\textbf{5.} Repeat \textbf{Step 2-4} until convergence
\end{itemize}
\caption{MC-PINNs for forward and inverse fractional PDEs}
\end{algorithm}

\section{Simulation results}
\label{S:4}
This section consists of 3 parts, which address the two types of data-driven problems set up in the introduction. We first investigate the performance of the MC-PINNs method for solving high-dimensional fractional Laplacian equations. Then we demonstrate the efficiency of the MC-PINNs method to solve inverse fractional advection-diffusion equations. Subsequently, we shall consider to solve a uncertainty quantification problem of FPDEs with unknown parameters.

In all our computations, we consider the $L_2$ relative error of the solution predicted by MC-PINNs:
\begin{equation}
\text{Relative}\ \  L_2 \ \ \text{error}= \frac{\|u_{NN}(x,t)-u(x,t)\|}{\|u(x,t)\|},
\end{equation}
where $u$ and $u_{NN}$ are the fabricated and surrogate solutions, respectively. We set $\epsilon=10^{-3}$ and $\epsilon_{t}=10^{-6}$ in the implementation for the stochastic approximation of the equation loss, and 1000 randomly chosen test points in the physical domain are chosen to compute the relative error. Unless stated otherwise, the DNNs model contains four hidden layers with 64 neurons per hidden layer.

\subsection{High-dimensional fractional Laplacian equation}
\label{S:4-1}
We start with the following fractional Laplacian equation
\begin{equation}\label{testFPDE}
  \left(-\Delta \right)^{\alpha/2} u \left(x \right)= f(x), \quad x\in
 \{x|\quad \|x\|_2^2\le 1 \}\subset \mathbb{R}^d,
\end{equation}
with zero boundary conditions. We consider a manufactured solution $u\left(x\right) = \left(1 -\| x \|_{2}^{2}  \right)^{1 + \alpha/2}$ and the corresponding forcing term is given by \cite{Dyda.2012}
\begin{equation*}
f\left(x\right) = 2^{\alpha} \Gamma\left(\frac{\alpha}{2}+2\right)\Gamma\left(\frac{\alpha+d}{2}\right)\Gamma\left(\frac{d}{2}\right)^{-1}\left(1 - \\ \left(1+\frac{\alpha}{d}\right) \| x \|_{2}^{2} \right).
\end{equation*}

We approximate $u(x,t)$ with $u_{NN}(x,t)=\text{Relu}(1-\|x\|^2)\tilde{u}_{NN}(x,t)$ in the simulations, where $\text{Relu}(z)=\max\{z, 0\}$ represents the rectified linear unit activation function. Thus we do not need to place training points on the boundary since $u_{NN}(x,t)$ satisfies the boundary conditions automatically. To ensure the best performance of the MC-PINNs, we use Adam optimizer with changing learning rate up to $M=10^4$ iterations. The number of residual points used for computing the equation loss for each mini-batch is taken as 128.

\begin{figure}[!ht]
\centering
\includegraphics[width=0.9\textwidth,height=0.20\textheight]{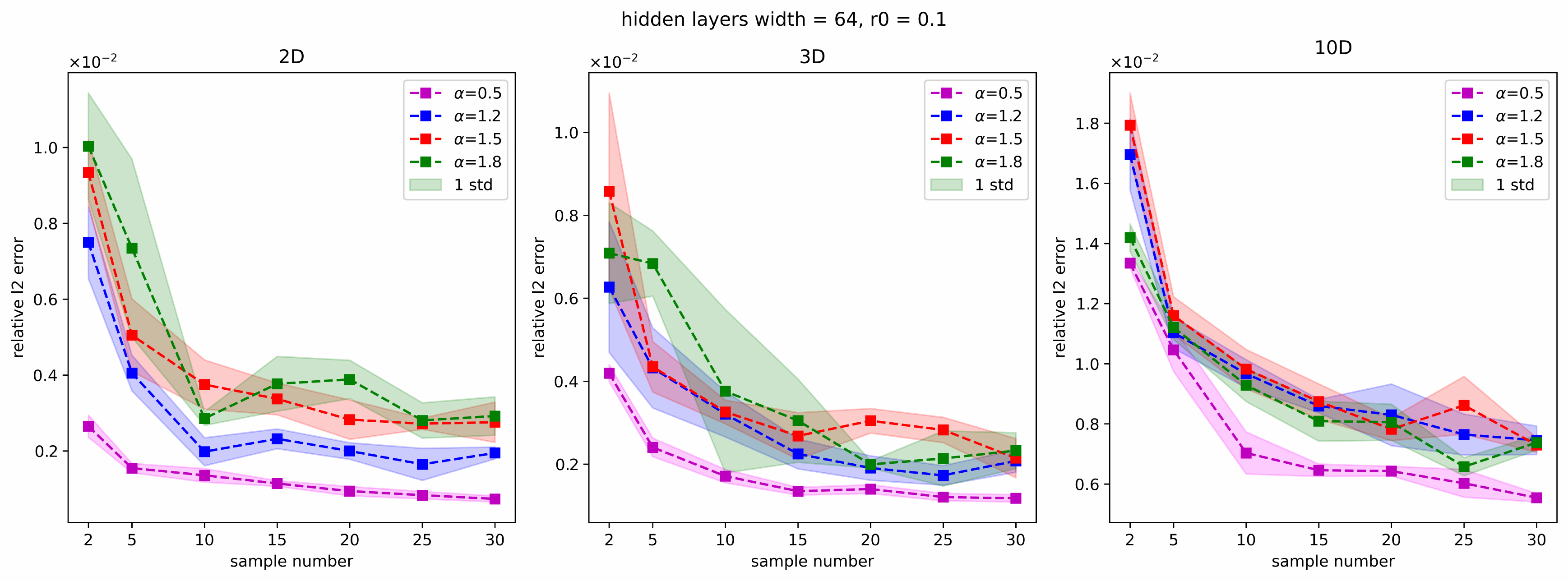}
\caption{Convergence for a fabricated solution $u\left(x\right) = \left(1 -\| x \|_{2}^{2}  \right)^{1 + \alpha/2}$. Relative $L^2$ error versus the parameter sample number for fixed number of $r_0$. Left: 2D; Middle: 3D; Right: 10D. The colored lines and shaded regions correspond to mean
values and one-standard-deviation bands of the MC-PINNs, respectively. \label{2D_r0}}
\end{figure}

\begin{figure}[!ht]
\centering
\includegraphics[width=0.9\textwidth,height=0.20\textheight]{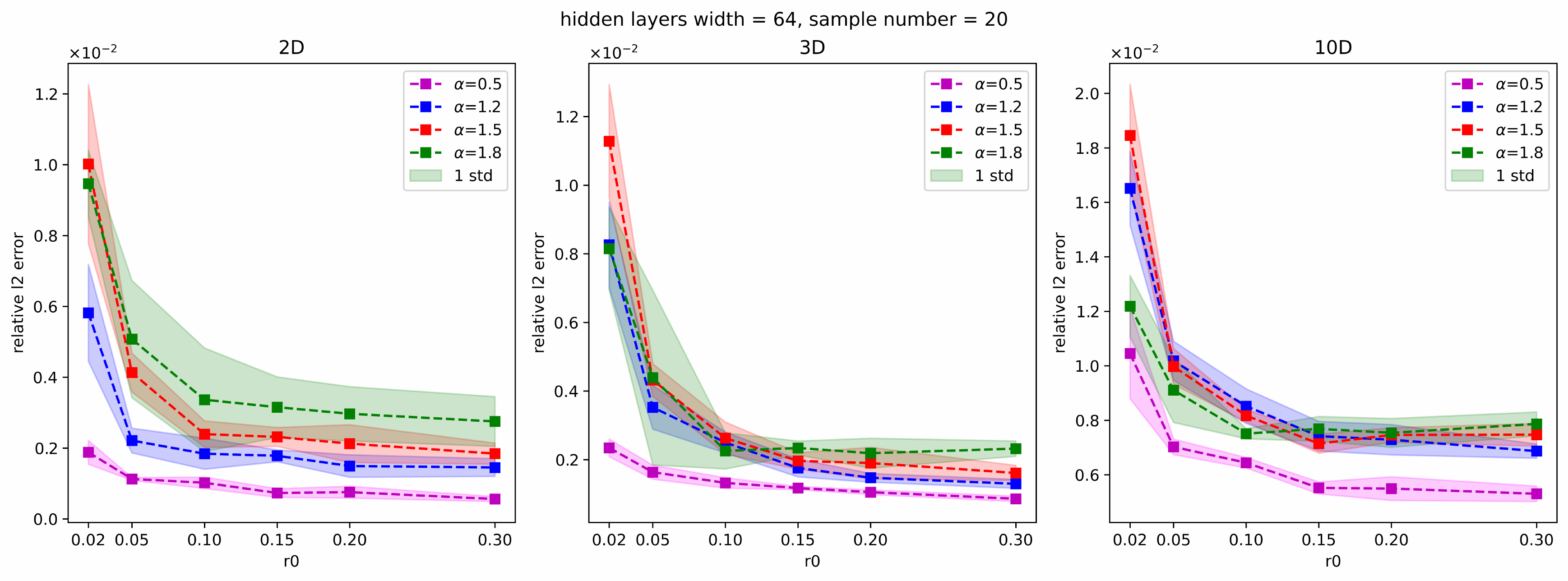}
\caption{Convergence for a fabricated solution $u\left(x\right) = \left(1 -\| x \|_{2}^{2}  \right)^{1 + \alpha/2}$. Relative $L^2$ error versus the parameter $r_0$ with fixed sample number 20. Left: 2D; Middle: 3D; Right: 10D. The colored lines and shaded regions correspond to mean
values and one-standard-deviation bands of the MC-PINNs solution errors, respectively.\label{2D_sample nmber}}
\end{figure}

We first report the impact of the sample number $m$ of random instrumental variables in (\ref{eqn:empericalloss}) and the neighborhood radius $r_0$ in (\ref{eqn:Flaplacian1}), which are used in the stochastic approximation of the fractional operators, on the accuracy of the MC-PINNs method for different fractional orders $\alpha=0.5$, $1.2$, $1.5$ and $1.8$ respectively. We run the MC-PINNs code five times for each fractional order and plot the mean and one standard-deviation band for the relative $L_2$ errors in Fig.~\ref{2D_r0} and Fig.~\ref{2D_sample nmber}. We can see that the error decays a little within a magnitude with increasing sample number when fixed $r_0$ and at the mean time the error is getting smaller as $r_0$ increasing with fixed sample number. We can also see that the uncertainty is decreasing for larger sample number and $r_0$. But finally the error saturates around $10^{-3}$, which show that the choice for the sample number and $r_0$ do not have significant impact on the accuracy.

\begin{figure}[!ht]
\centering
\includegraphics[width=0.25\textwidth]{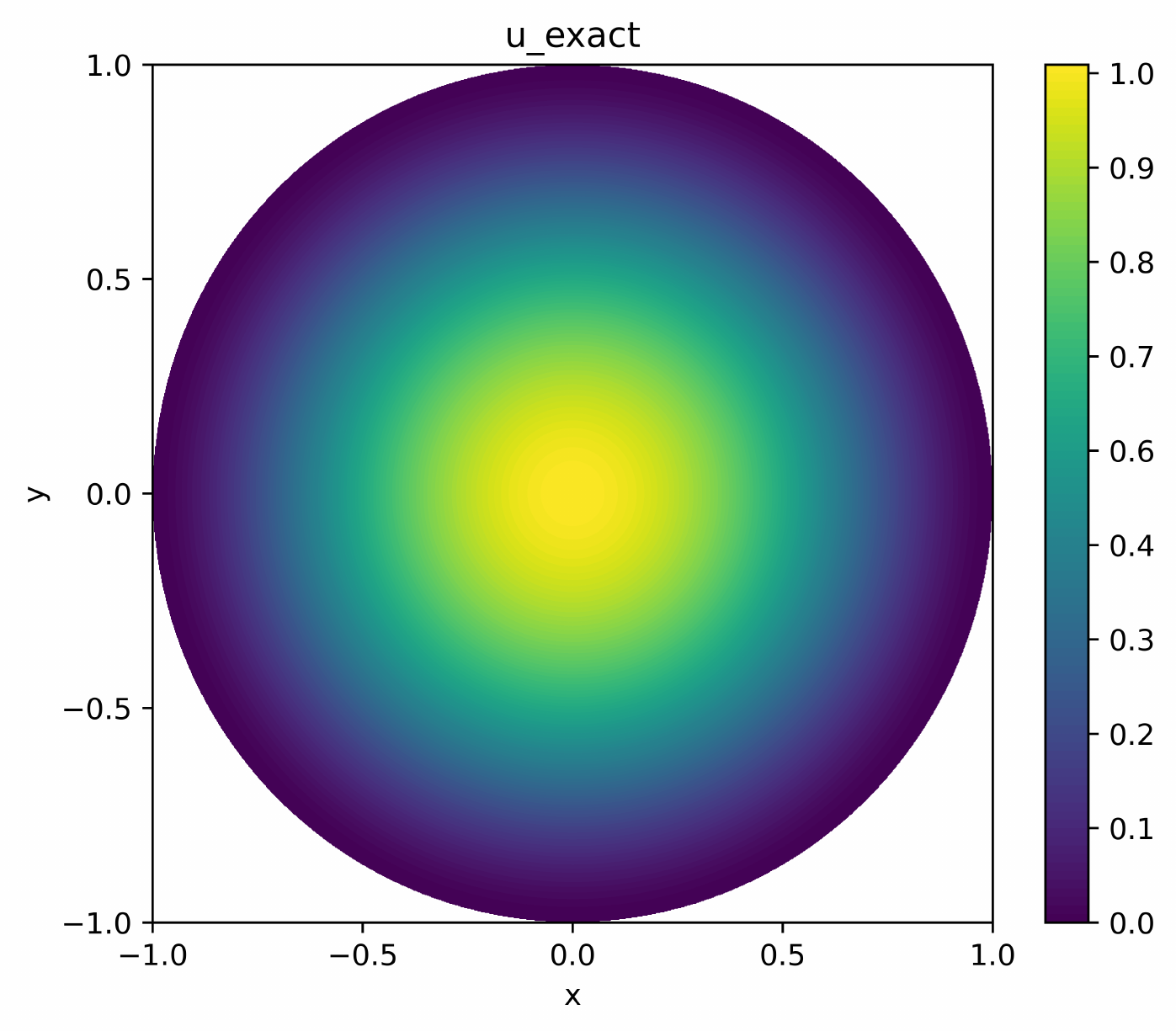}
\includegraphics[width=0.25\textwidth]{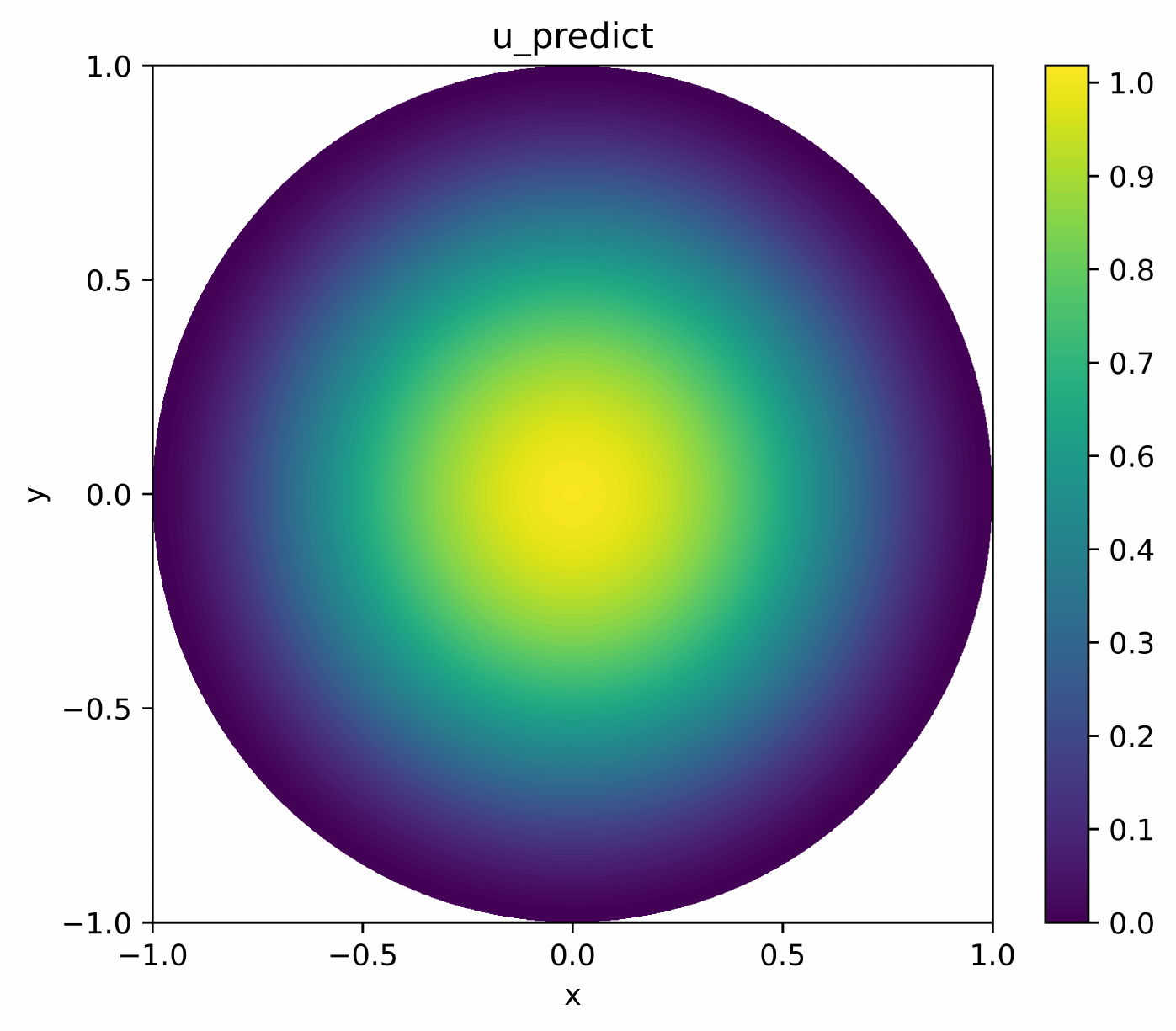}
\includegraphics[width=0.25\textwidth]{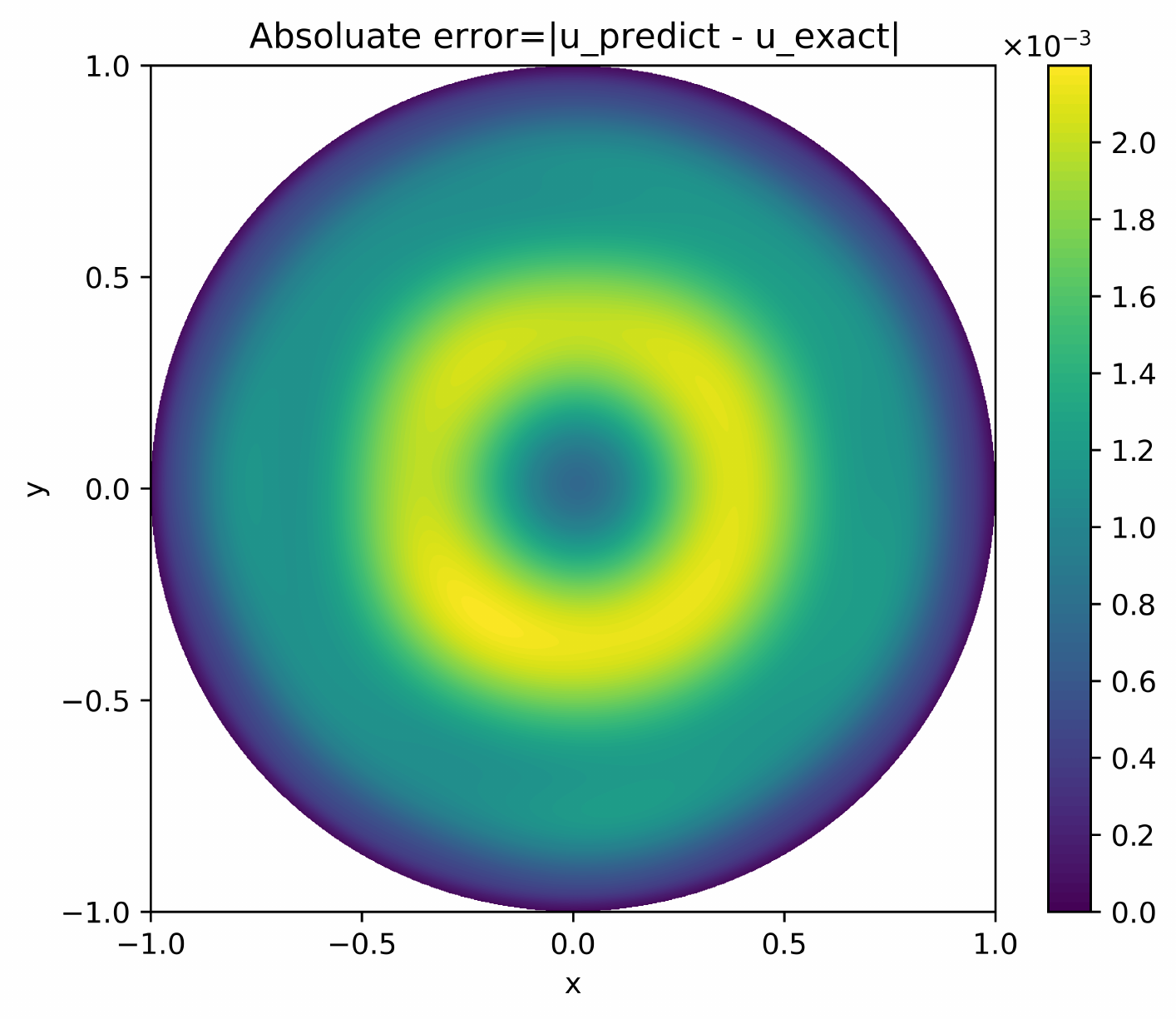}
\includegraphics[width=0.25\textwidth]{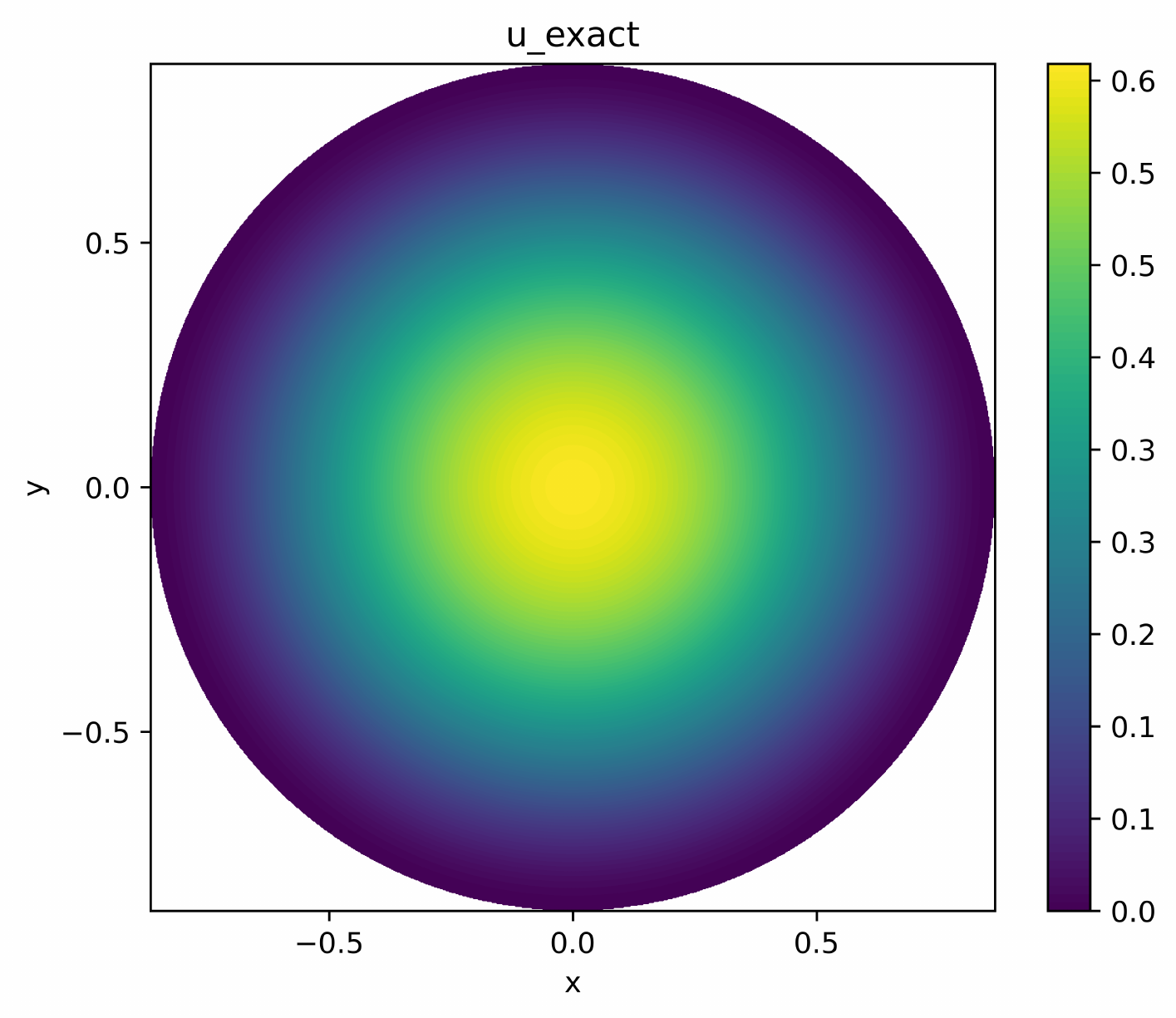}
\includegraphics[width=0.25\textwidth]{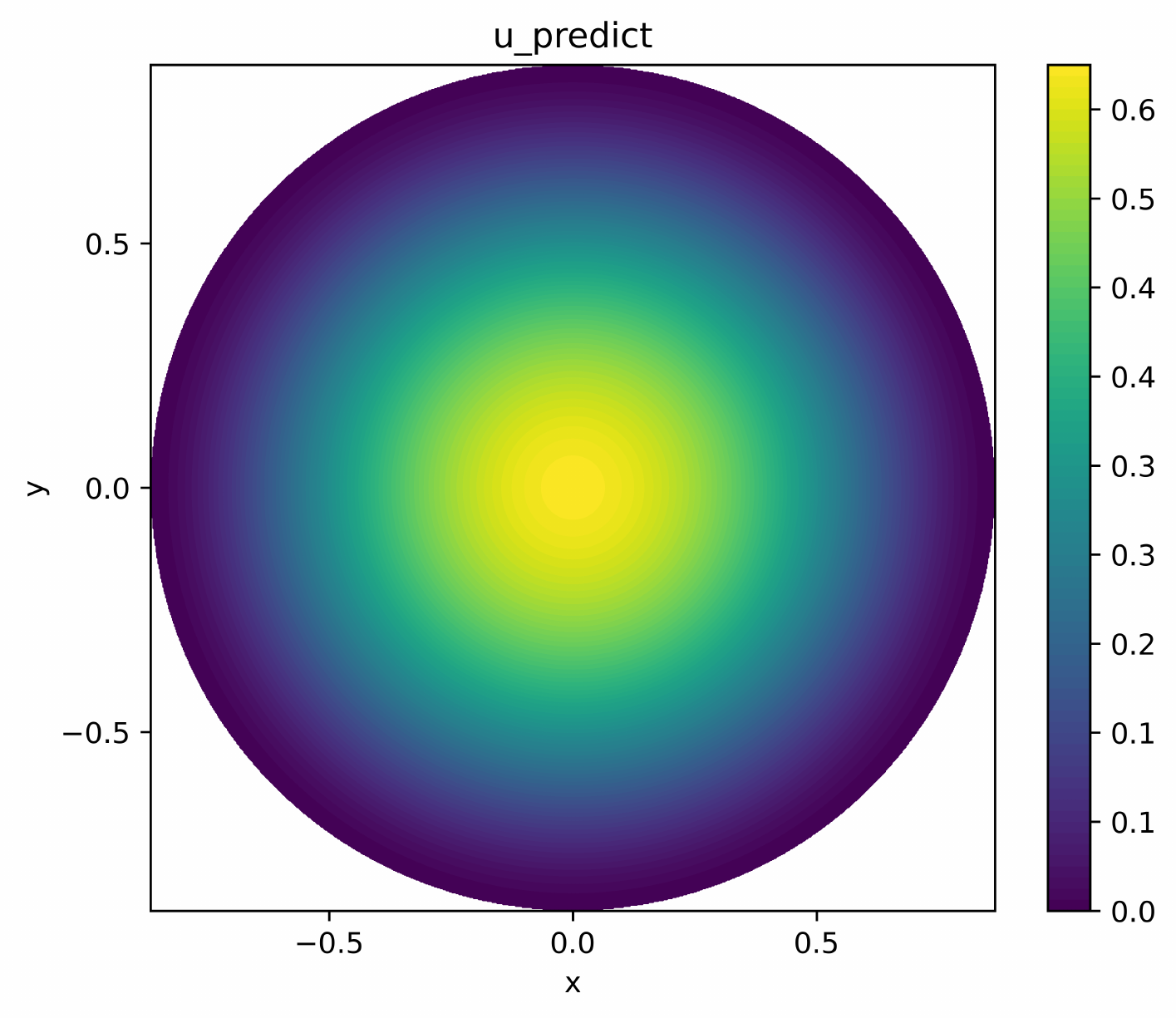}
\includegraphics[width=0.25\textwidth]{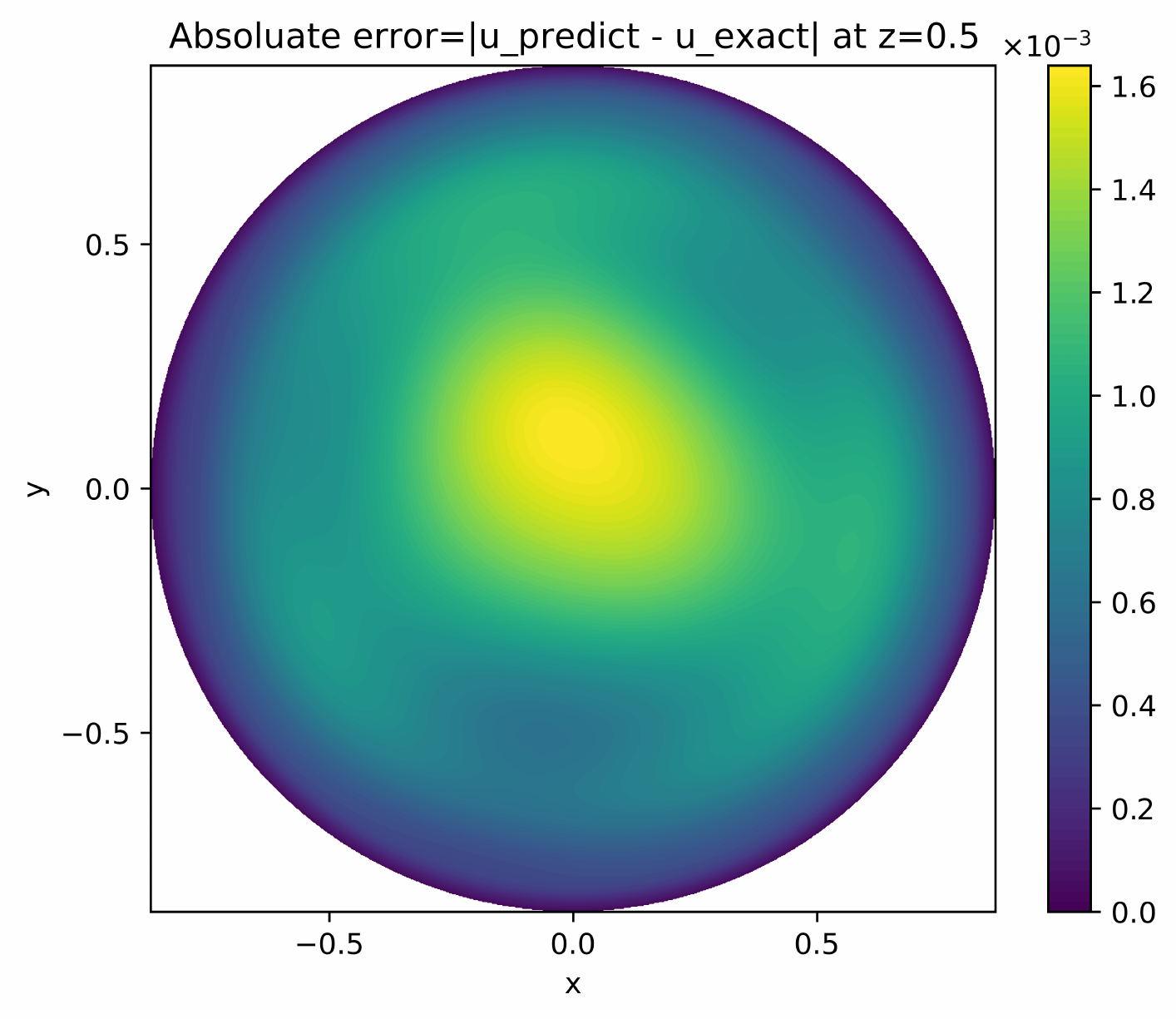}
\caption{MC-PINNs accuracy in multidimensional simulations of fractional Laplacian with
the fabricated solutions $u\left(x\right) = \left(1 -\| x \|_{2}^{2}  \right)^{1 + \alpha/2}$. \text\bf{Top:} 2D exact solution (left), the MC-PINNs solution (middle) and the corresponding absolute error (right). \text\bf{Bottom:} 3D exact solution (left), the MC-PINNs solution (middle) and the corresponding absolute error (right). \label{fPDE_accuracy}}
\end{figure}

Based on the above results on the relative error depending on $r_0$ and the sample number, next we show the accuracy of the MC-PINNs method for 2D, 3D and 10D fractional Laplacians with $\alpha=1.5$, respectively. We run the MC-PINNs code five times with fixed sample number 25 and $r_0=0.2$ and plot the mean prediction. Fig.~\ref{fPDE_accuracy} displays the contour plots of the fabricated solutions, MC-PINNs predict solutions, and the absolute errors of the solutions in comparison with the fabricated solutions, respectively. We can see that the error is around $10^{-3}$, which are sufficiently low especially for the high-dimensional problem.

%

Finally, we report the computational cost and flexibility of the MC-PINNs compared with fPINNs proposed in \cite{pang2019fpinns}. To approximate the time-fractional derivative of the neural network output $u_{NN}(x,t),$ a finite difference $L_1$ scheme is employed in \cite{pang2019fpinns}. Concretely, for a fixed training point $(x,t)$, evaluations of some auxiliary points are necessary to compute the time fractional derivative at this point using the $L_1$ scheme due to the nonlocal property of the fractional derivative. This procedure makes the discretization of the fractional Laplacian operator in space more complicated. For example, to cope with the fractional Laplacian defined in the sense of directional derivatives, the shifted vector Gr\"uwald-Letnikov (GL) formula is first adopted and then quadrature rule is employed to approximate
$\int_{0}^{2\pi}(\cdot)d\vartheta$ in $(-\Delta)^{\alpha/2}$. We use the DeepXDE package developed in \cite{lu2021deepxde} to compute the times that we need to calculate the fractional Laplacian of $u_{NN}$ at a fixed training point during the training stage. For the second-order GL scheme,
we need to calculate $u_{NN}$ around $q^{d}\times L$ times
to approximate the fractional Laplacian at a given $x$
by selecting $q$ quadrature points for each angular coordinate and $L$ auxiliary points for each quadrature point, where $d$ is the dimension of physical space.
In \cite{pang2019fpinns}, $q\ge 8$ and the average $L$ is about $100$. While by using the Monte Carlo calculation of the fractional Laplacian we can obtain the number we only need to calculate $u_{NN}$ around $8\times m+1$ times, here $m$ is the sample number we used to do the MC approximation and was set to be $20$ in our simulations. Thus we conclude that MC-PINNs alleviate the computational cost greatly compared with fPINNs, which set up the potential for MC-PINNs to solve 10D problems.

\subsection{Inverse problems of fractional advection-diffusion equation (ADE)}
We now present the performance of using MC-PINNs to solve an inverse problem for the space-time fractional differential equation defined by Eq.~(\ref{eqn:SDE}). Specifically, we consider a fabricated solution $u(x,t)=(1-\|x\|^2)e^{-t}$ \cite{pang2019fpinns}. According to \cite{Dyda.2012} and \cite{gorenflo2020mittag}, the space-fractional and the time-fractional derivatives of $u(x,t)$ can be computed analytically and thus we can obtain the forcing term.
\begin{equation}\label{eqn:adef}
\begin{aligned}
f\left(x,t\right)& =-t^{1-\gamma}E_{1,2-\gamma}(-t)(1-\|x\|^2)^{1+\frac{\alpha}{2}} \\
&   +c2^{\alpha} \Gamma\left(\frac{\alpha}{2}+2\right)\Gamma\left(\frac{\alpha+d}{2}\right)\Gamma\left(\frac{d}{2}\right)^{-1}\left(1 - \left(1+\frac{\alpha}{d}\right) \| x \|_{2}^{2} \right)e^{-t} +(1+\frac{\alpha}{2}(1-\|x\|^2)^{\frac{\alpha}{2}}(-2x))e^{-t},
\end{aligned}
\end{equation}
where $E_{a,b}(t)$ the Mittage--Leffler function defined by $E_{a,b}(t)=\sum\limits_{k=0}^{\infty}\frac{t^k}{\Gamma(ak+b)}$.

Solving the inverse problem with MC-PINNs has the same flowchart as the forward problem without changing any code, the only thing we need to do is to let the FPDE parameters, which are the targets to be identified, to be optimized together with the DNN parameters during the training process.  We now assume that we do not know the exact fractional orders $\alpha$ and $\gamma$, the diffusion coefficient $c$ and the flow velocity $v$. We want to use the MC-PINNs method to identify these unknown parameters and the solution $u(x,t)$. Extra measurements from $u$ could help us infer these coefficients. In the domain $\Omega\times 
\{t=T\}$, we select $N_u=20, 80, 100$ additional uniformly distributed measurements of $u$ for the 1D/3D/5D problems, respectively.

The "hidden" values of $\alpha$, $\gamma$, $c$ are selected to be $1.5$, $0.5$ and $0.1$, respectively. The true value of $v$ will be specified for 1D/3D/5D problems in Table \ref{tb1-ADE}. When setting up the MC-PINNs, the unknown parameters are coded as "variables" instead of as "constants" so that they will be tuned at the training stage. Without loss of generality, the initial values of the unknown parameters are taken as $\alpha_0=1.7$, $\gamma=0.9$, $c=0.5$. The initial value for $v$ is taken from $U[0,0.1]^d$, where $U[0,0.1]$ is uniform distribution within $[0,0.1]$. In practice, these values could be chosen based on reasonable guesses. The sample number $m$ of random instrumental variables in (19) is set to be 30 and the neighborhood radius $r_0=0.3$. We use 128 residual points for computing the equation loss for each mini-batch. The neural networks are trained with an Adam optimizer with changing leaning rate for 40000 epochs.

\begin{figure}[!ht]
\centering
\includegraphics[width=0.3\textwidth,height=0.20\textheight]{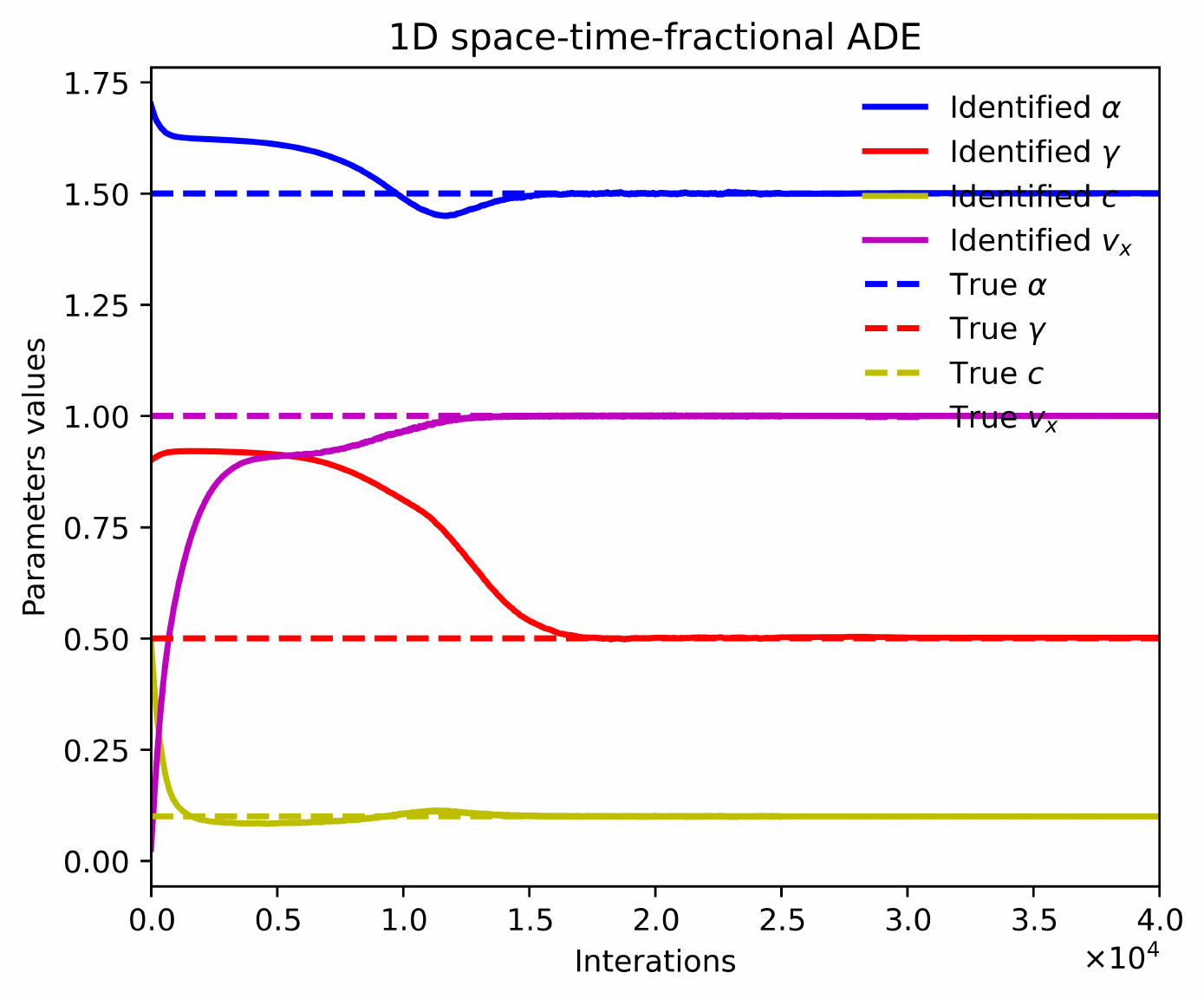}
\includegraphics[width=0.3\textwidth,height=0.20\textheight]{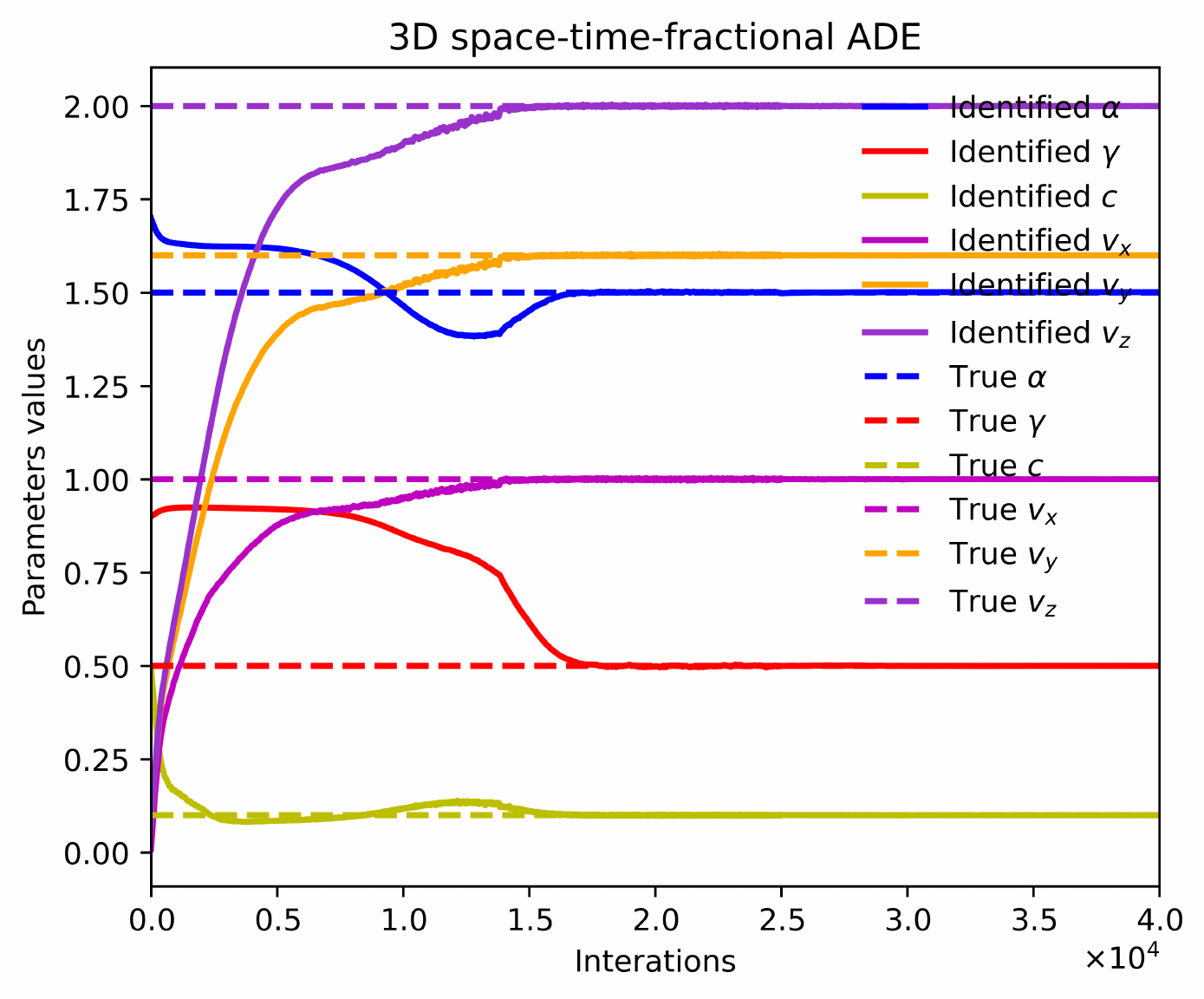}
\includegraphics[width=0.3\textwidth,height=0.20\textheight]{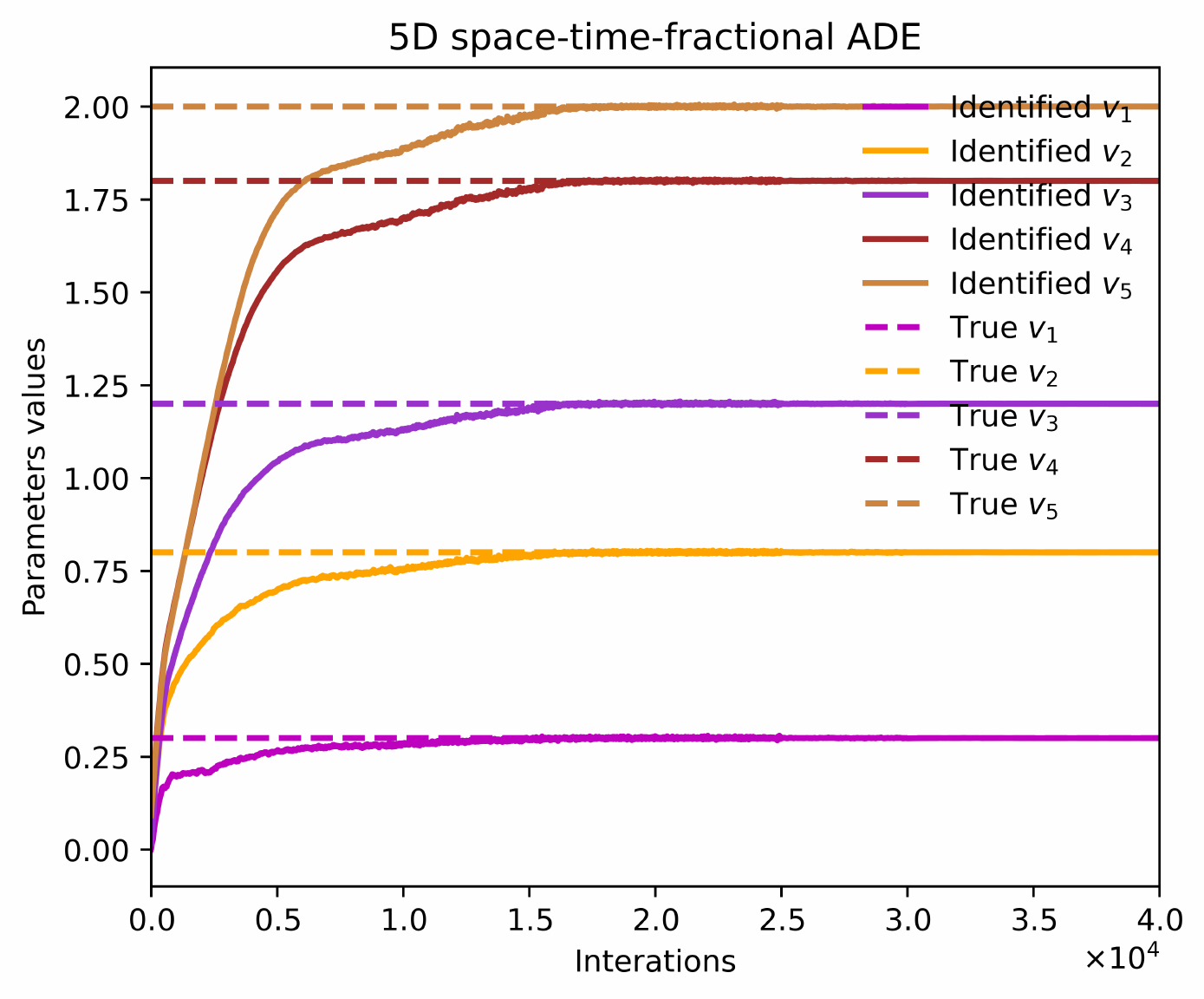}
\caption{ADE Inverse problem for parameters identification. Parameter evolution as
the iteration of optimizer progresses: \text{Left}: 1D space-time-fractional ADE; \text{Middle}: 3D
space-time-fractional ADE; and \text{Right}: 5D space-time-fractional ADE. \label{ADE_parameters}}
\end{figure}

\begin{figure}[!ht]
\centering
\includegraphics[width=0.25\textwidth]{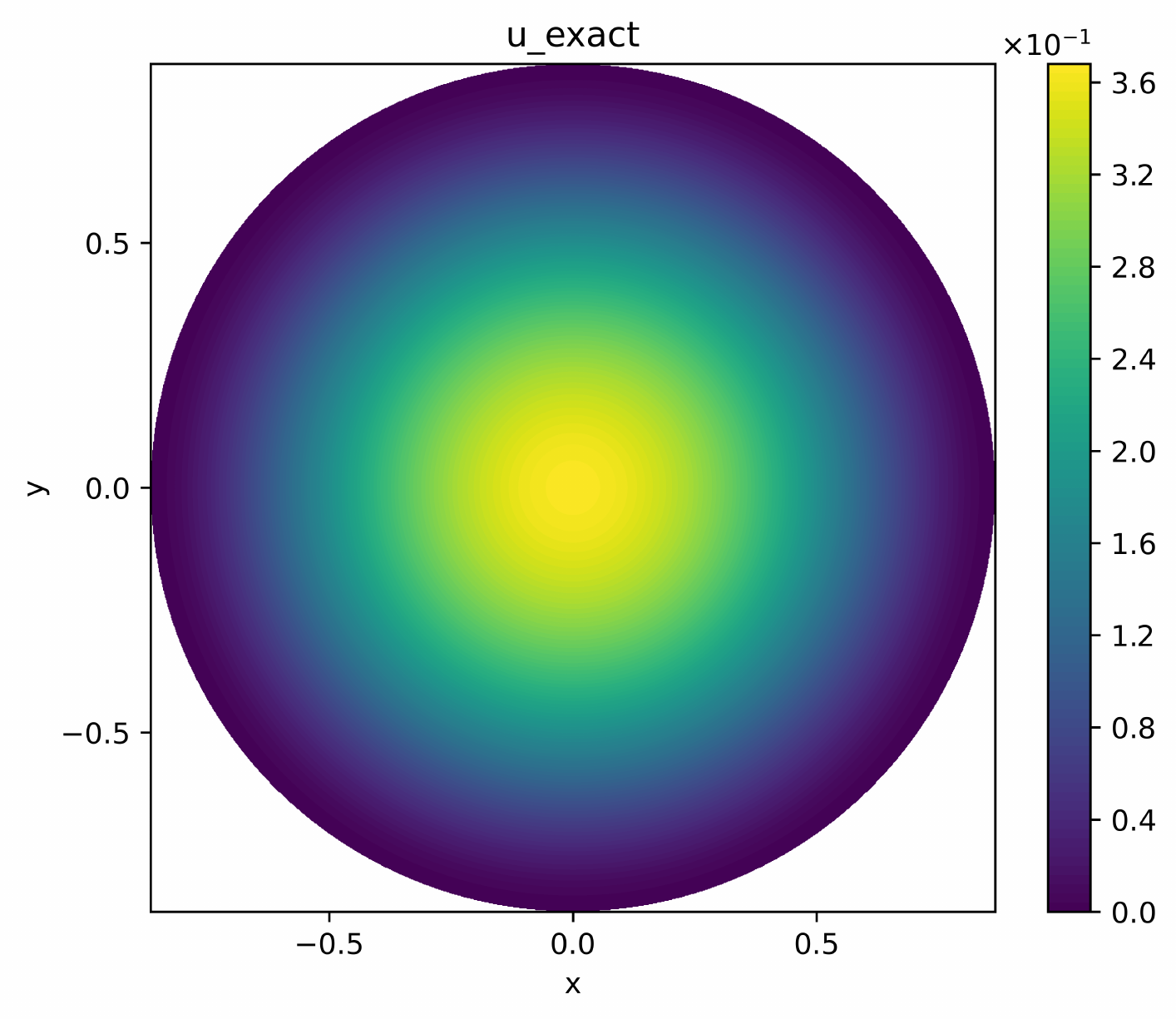}
\includegraphics[width=0.25\textwidth]{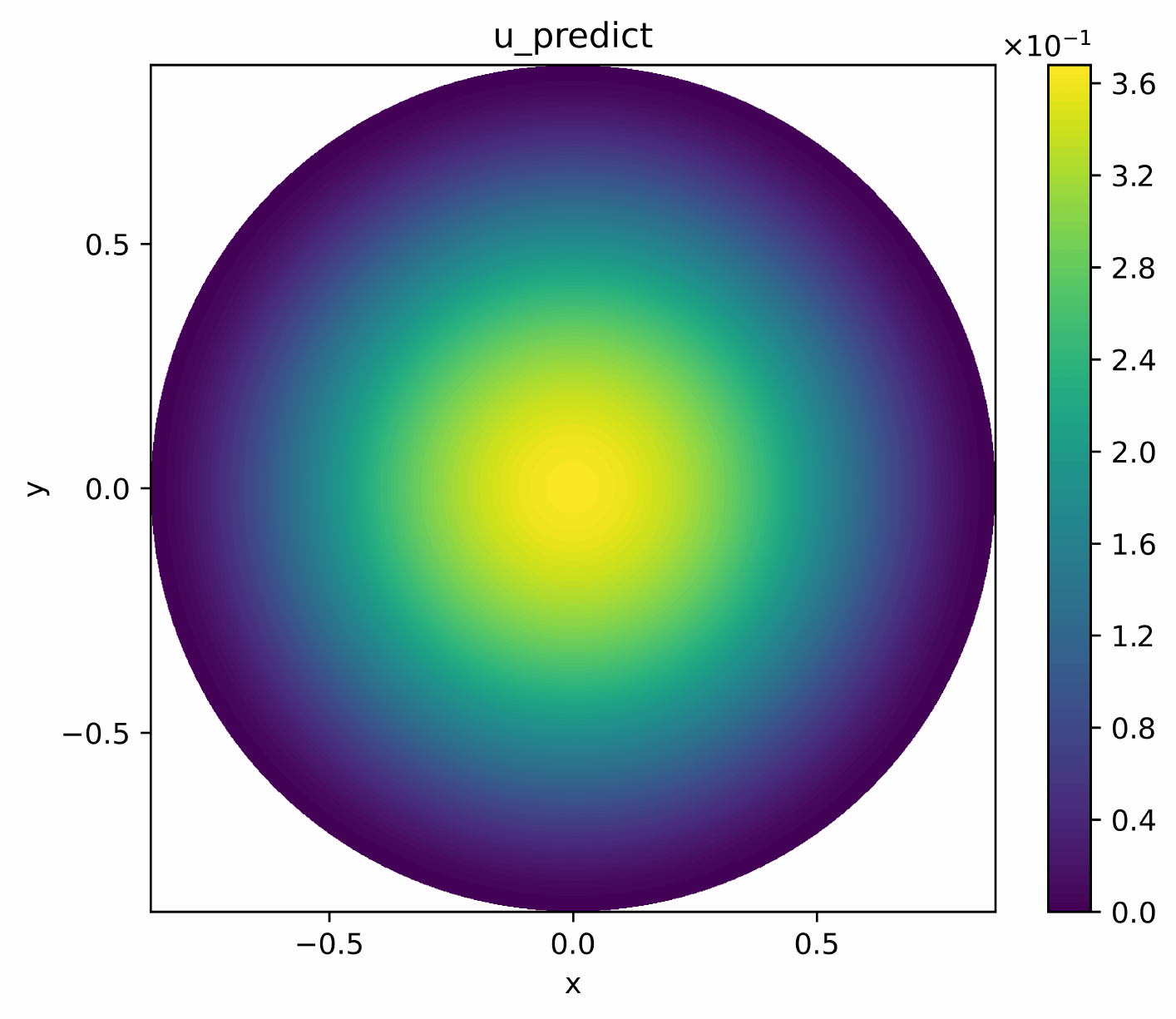}
\includegraphics[width=0.25\textwidth]{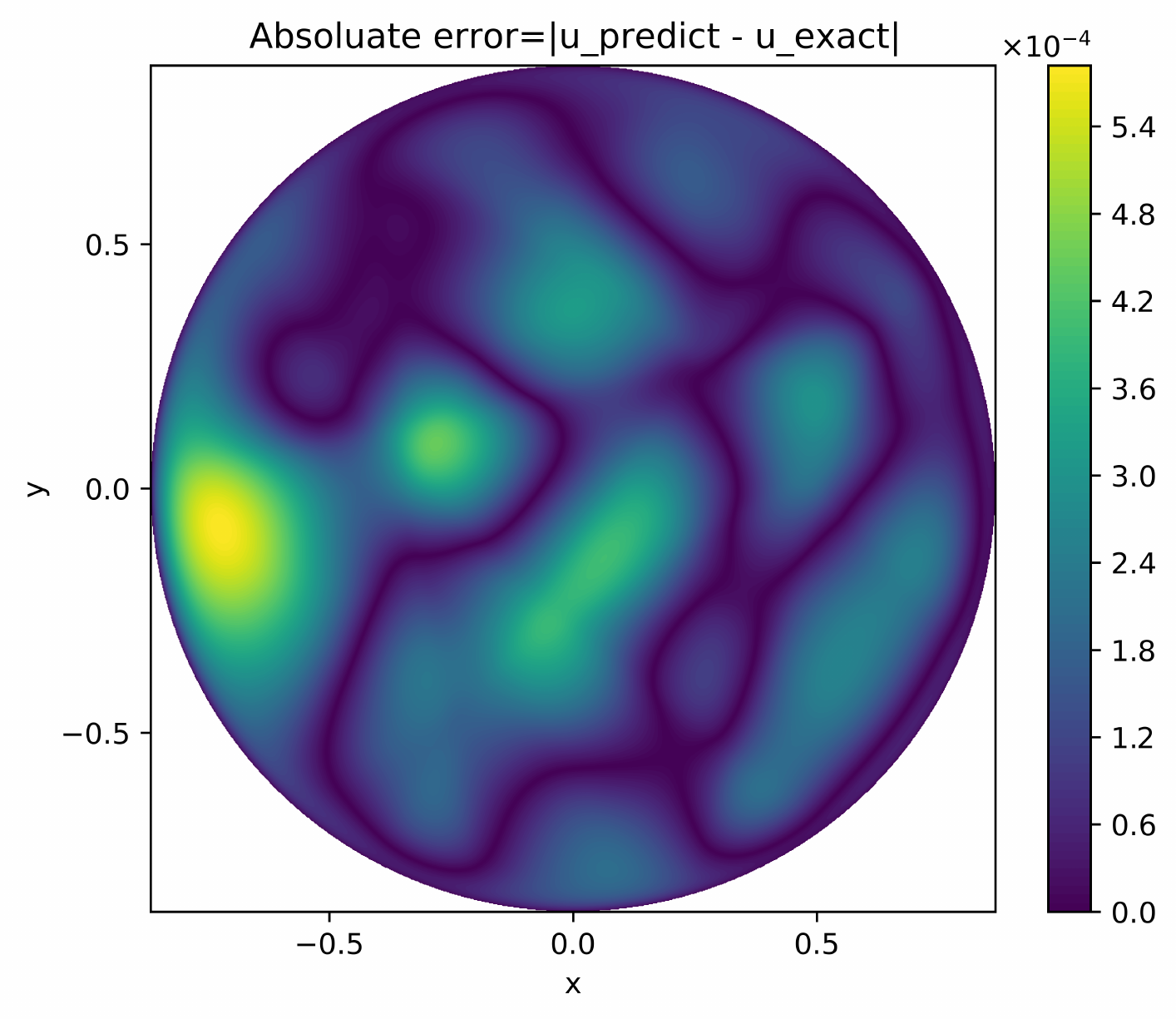}
\includegraphics[width=0.25\textwidth]{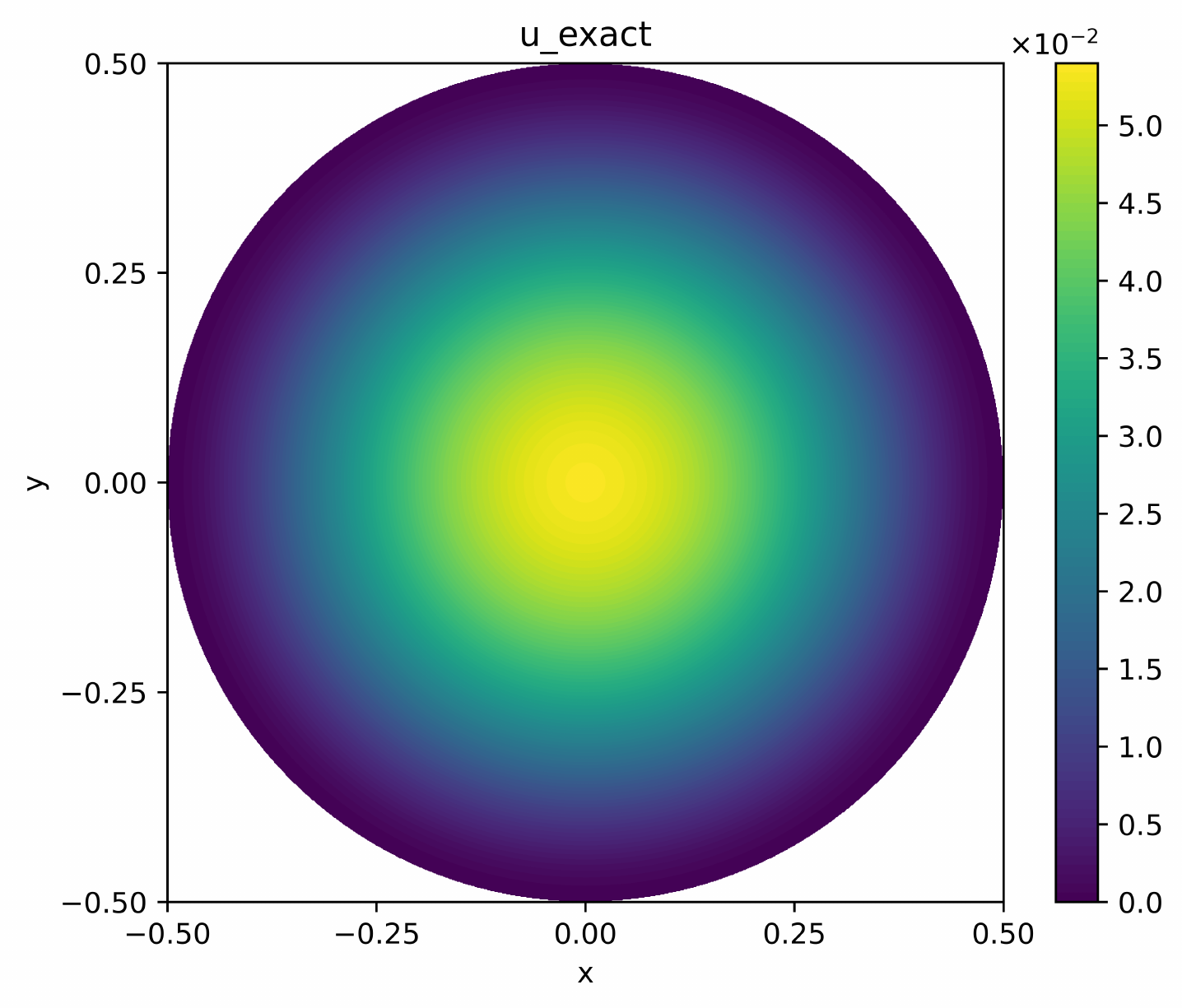}
\includegraphics[width=0.25\textwidth]{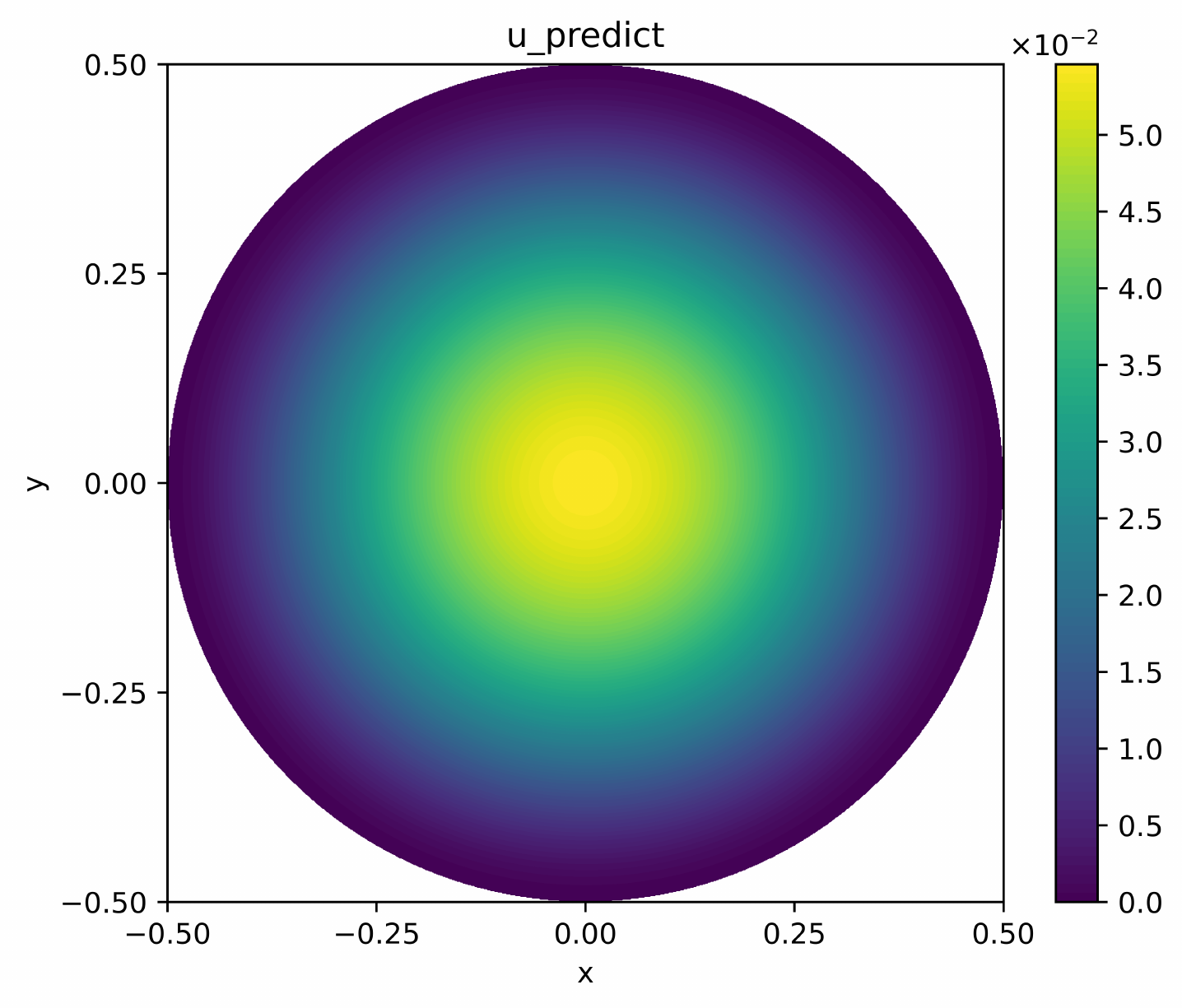}
\includegraphics[width=0.25\textwidth]{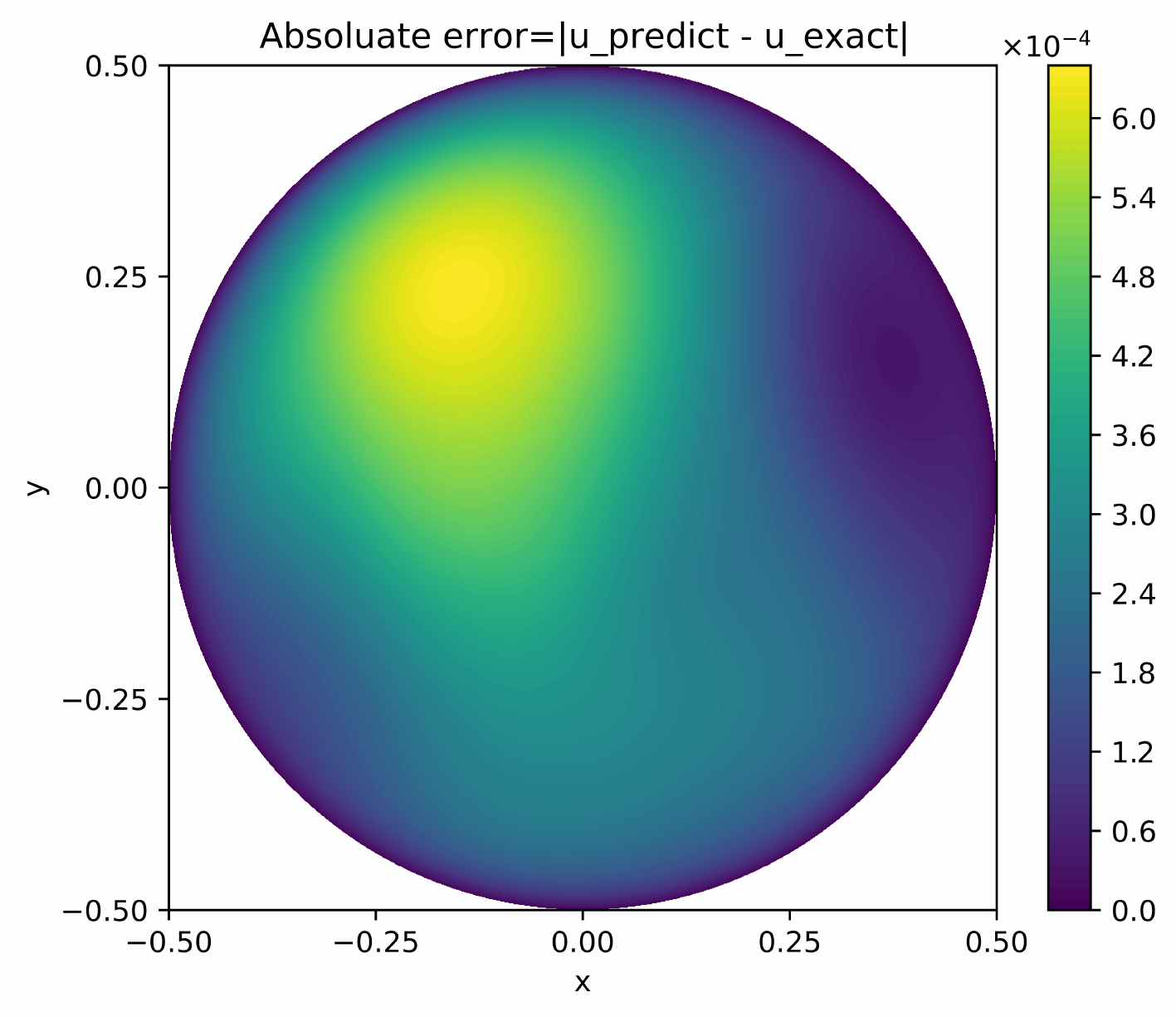}
\caption{MC-PINNs recovery accuracy for inverse space-time-fractional ADE. \text\bf{Top:} 3D exact solution (left), the MC-PINNs recovered solution (middle) and the corresponding absolute error (right). \text\bf{Bottom:} 5D exact solution (left), the MC-PINNs recovered solution (middle) and the corresponding absolute error (right).\label{ADE_solution}}
\end{figure}

Fig.~\ref{ADE_parameters} displays the convergence history of the parameters for the 1D/3D/5D fractional ADEs. We can observe that the inferred values converge to the true values after less than 20000 training epochs for all the problems. Fig.~\ref{ADE_solution} shows the contour plots of the fabricated solutions, MC-PINNs recovered solutions, and the absolute
errors of the recovered solutions in comparison with the fabricated solutions, respectively. Table \ref{tb1-ADE} is a simulation summary of the above parameter estimation and also we listed the recovered parameters and the relative $L_2$ error for 1D/3D/5D problems. We can observe that all the hidden FPDE parameters and the solution field $u$ are well identified. And the error becomes deteriorate as the physical dimension goes higher.

\begin{table}[ht!]
\linespread{1}
\footnotesize
\caption{\footnotesize Identified parameters and relative errors of the predicted solution field $u$ for 1D/3D/5D inverse
ADE problems with synthetic data. The true parameters are $\alpha  = 1.5$, $\gamma  = 0.5$, $c =0.1$. }\label{tb1-ADE}
\begin{center}
\begin{tabular}{|c|c|c|c|c|c|c|c|c|c|c|}
\hline
 & \text{True parameters}($\alpha$, $\gamma$, $c$, $v$)& \text{Identified parameters}($\alpha$, $\gamma$, $c$, $v$) & \text{Relative} $L_2$ error  \\
\hline
$1D$& $1.5,0.5,0.1,1.0$   & $1.50060,0.50181,0.09993,0.99997$  & $5.04\times 10^{-4}$    \\
\hline
$3D$& $1.5,0.5,0.1,1.0,1.6,2.0$   & $1.50115,0.50005,0.09981,1.00000,1.59993,2.00000$   &$1.18\times 10^{-3}$     \\
\hline
$5D$& $1.5,0.5,0.1,0.3,0.8,1.2,1.8,2.0$   & $1.50188,0.49736,0.09968,0.29976,0.79998,1.20005,1.80015,2.00016$ &$3.26\times 10^{-3}$      \\
\hline
\end{tabular}
\end{center}
\end{table}

\subsection{Fractional diffusion equation with random inputs}
Finally, we consider the following parametric diffusion equation with fractional Laplacian
\begin{equation}\label{eqn:FDE}
\begin{array}{rcl}
(-\Delta)^\frac{\alpha}{2}u(x)+\mu u(x) & = & f(x,\alpha_0),\quad x\in \Omega\subset\mathbb{R}^d, \\
u(x) & = & 0, \quad \qquad x\in \mathbb{R}^{d}\backslash\Omega,
\end{array}
\end{equation}
where $\alpha$ and $\mu$ are input random variables in this example, and $\Omega=\{x|\Vert x\Vert\le 1\}$. Specifically, we assume $\alpha\sim U[0.5,1.5]$, $\mu\sim U[-0.5,0.5]$.  $\alpha_0=1$ is a constant in the forcing term. Our goal for this simulation is to build a DNN surrogate for $u(x|\alpha,\mu)$ given any $\alpha$ and $\mu$. Thus we can get the statistical approximation of $u$ after the parameters in DNN are fine tuned. The number of residual points used for computing the equation loss for each mini-batch is taken as 128. The sample number $m$ is set to be 30 and the neighborhood radius is $r_0=0.2$. The neural networks are trained with an Adam optimizer with changing leaning rate for 10000 epochs.

\begin{figure}[!ht]
\centering
\includegraphics[width=0.3\textwidth,height=0.2\textheight]{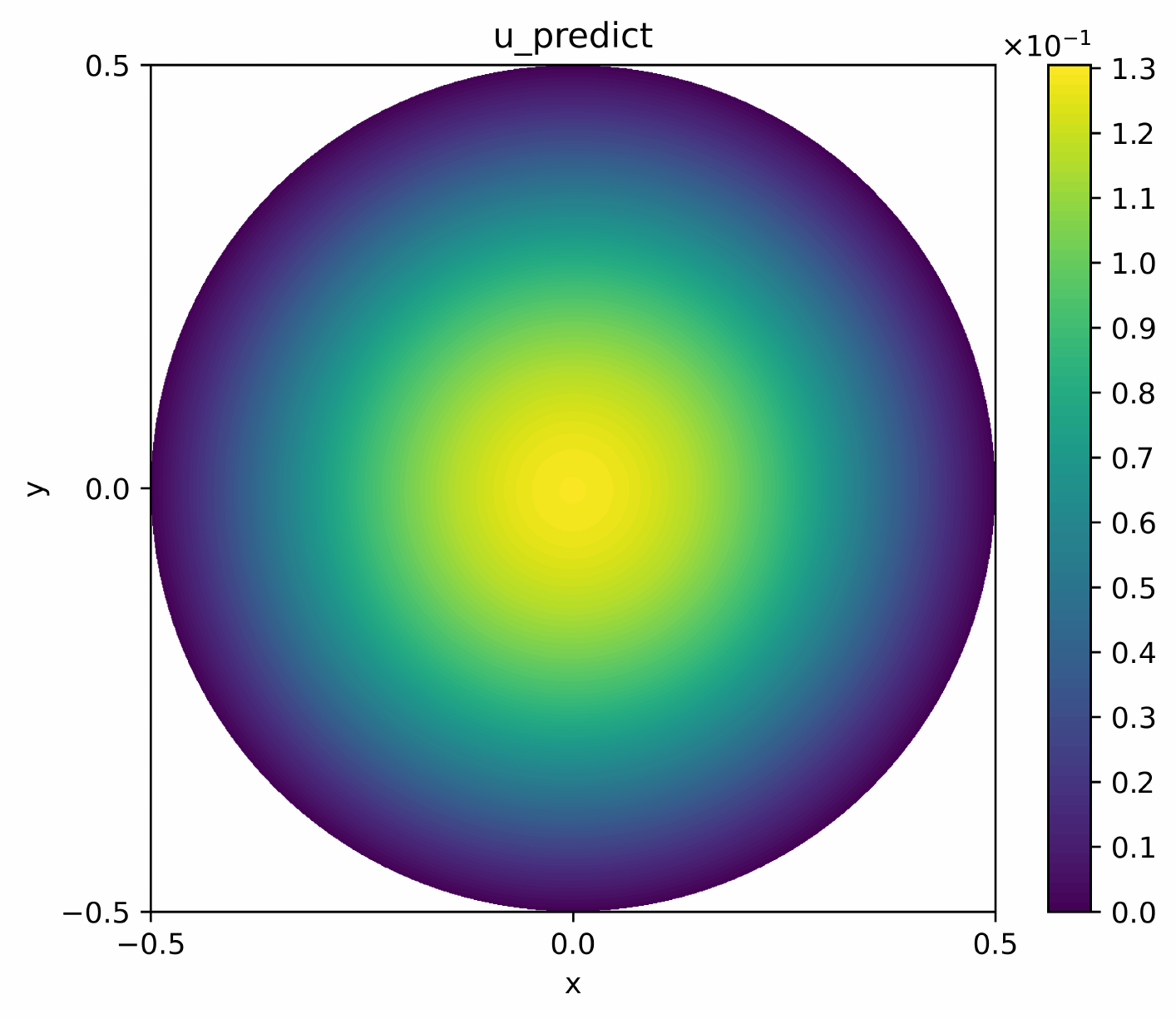}
\includegraphics[width=0.3\textwidth,height=0.2\textheight]{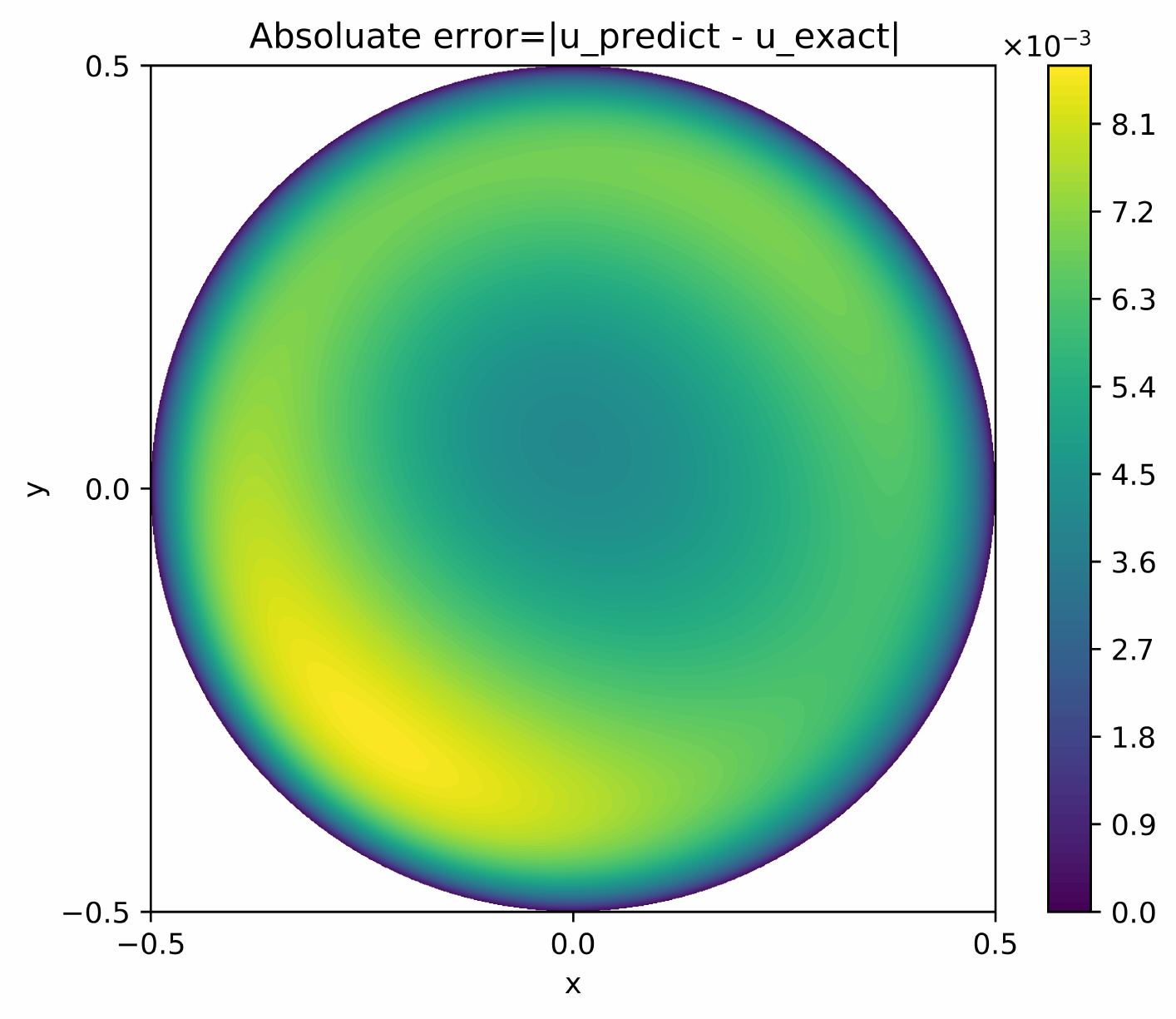}
\includegraphics[width=0.3\textwidth,height=0.2\textheight]{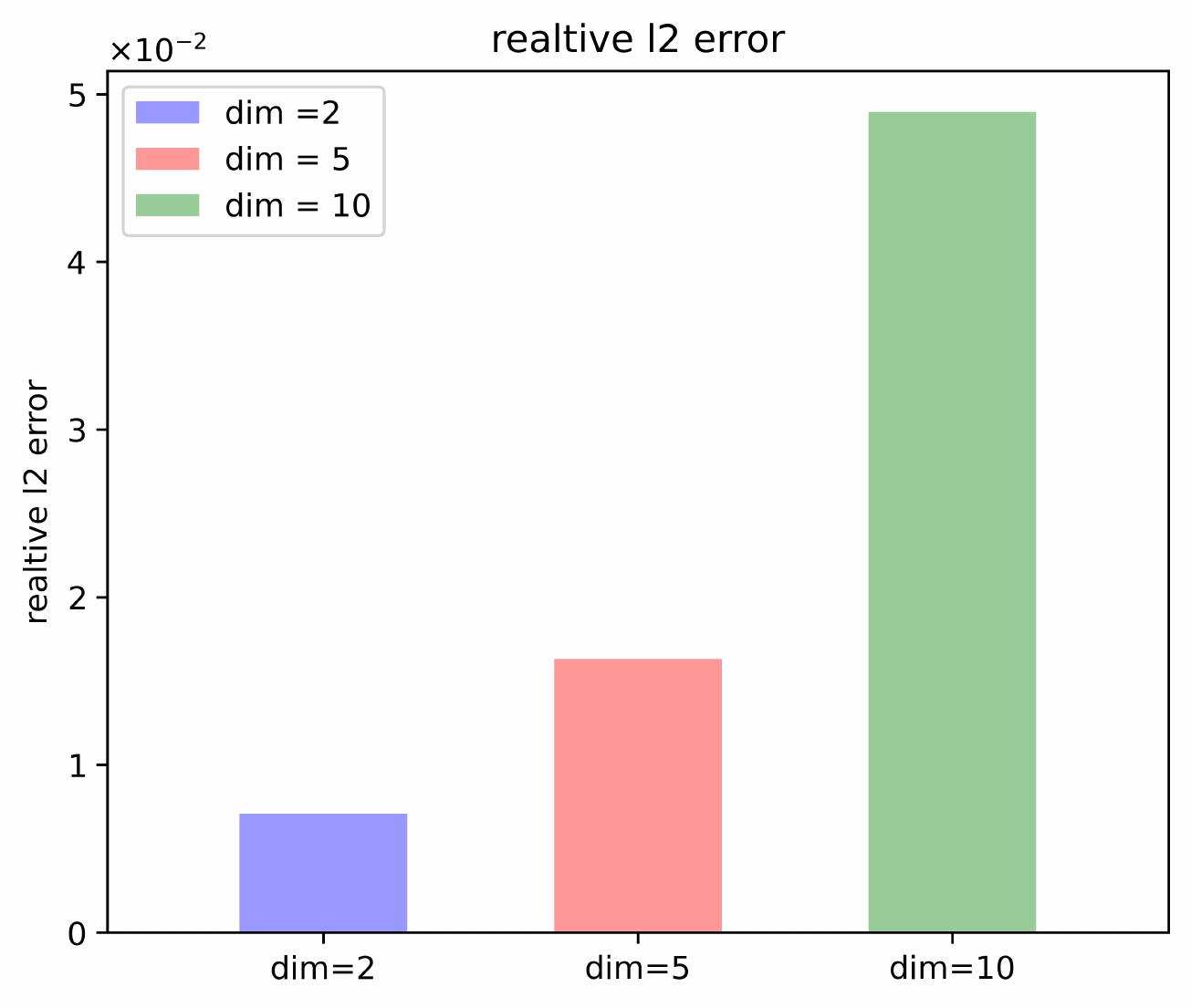}
\caption{Fractional diffusion equation with random inputs. \text{Left:} contour of the MC-PINNs recovered solution in 5D; \text{Middle:} contour of the absolute error between the MC-PINNs recovered solution and the fabricated solution in 5D; \text{Right:} Relative $L_2$ error for 1D,5D and 10D respectively. \label{random_error}}
\end{figure}

We will demonstrate the performance of the MC-PINNs method for 2D/5D/10D problems. We investigate the accuracy of the MC-PINNs solution with fixed $\alpha_0=1$,
$\alpha=\alpha_0$ and $\mu=0$. This is a special case for the problem considered in Section 4.1. The left and middle plots of Fig.~\ref{random_error} show the contour plot of the MC-PINNs prediction and the absolute
errors of the predicted solutions in comparison with the fabricated solutions for the 5D problem respectively. The right plot of Fig.~\ref{random_error} plots the relative $L_2$ error for 2D/5D/10D problems. We can see that the error is increasing for high-dimensional problem since the same number of residual points are used during the training process. 

\begin{figure}[!ht]
\centering
\includegraphics[width=0.4\textwidth,height=0.25\textheight]{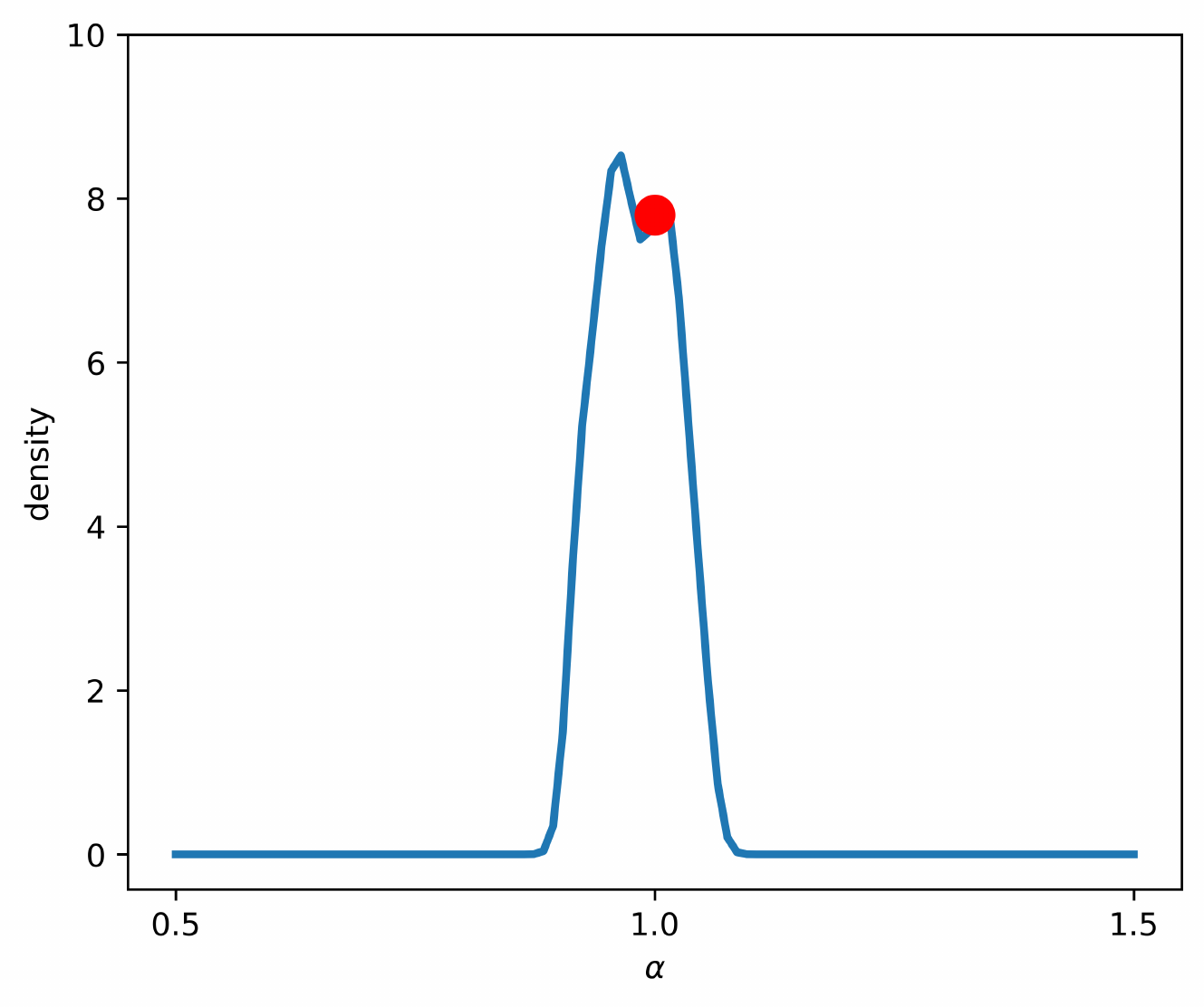}
\includegraphics[width=0.4\textwidth,height=0.25\textheight]{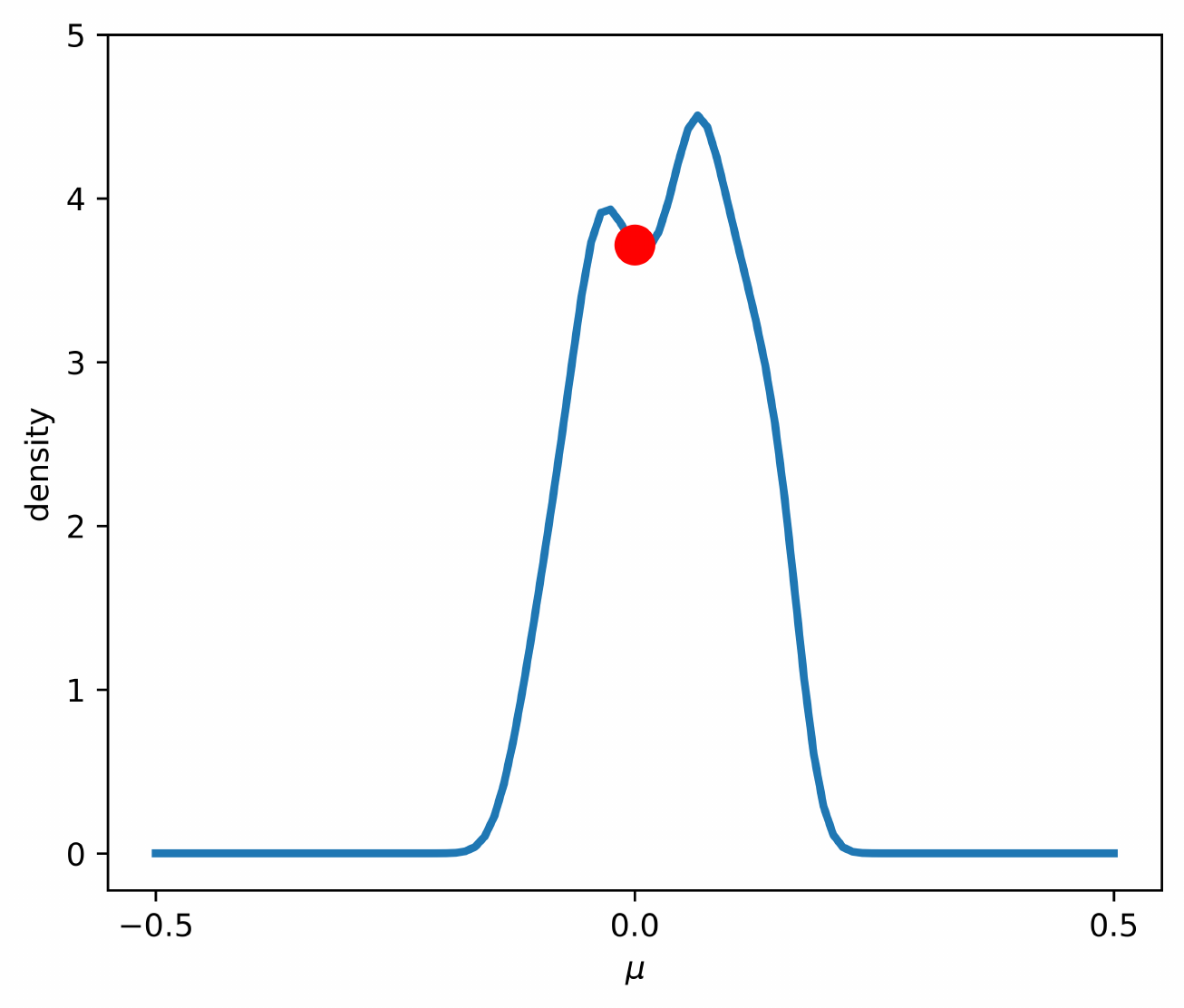}
\caption{Approximate posterior distributions of parameters $\alpha$ and $\mu$ for $d=2$, where the sensor positions are $(-0.0120, -0.2170)$, $(0.0321,  0.7628)$, $(0.6677,  0.1095)$, $(0.5411,  0.5840)$, $(-0.2382, -0.7787)$. The density values are calculated by the \texttt{gaussian\_kde} function of \texttt{scipy.stats} module in Python, and red points represent true values of parameters $\alpha=1,\mu=0$. \label{random_abc}}
\end{figure}

The surrogate model given by MC-PINNs can also be applied to parameter identification of the diffusion equation.
As an example, we assume that $u$ is defined as in Section \ref{S:4-1} for $d=2$, parameters $\alpha,\mu$ are unknown,
and $5$ sensors for $u$ are placed in $\Omega$.
Then, we can use the approximate Bayesian computation method to calculate the posterior distribution of parameters according to histograms of
\[
\left\{(\alpha_{i},\mu_{i})|\sum_{k=1}^{5}\left(u(x_{k}|\alpha_{i},\mu_{i})-u_{k}\right)^2\le\epsilon_{\mathrm{abc}},1\le i\le N_{\mathrm{abc}}\right\}
\]
where $x_k,u_k$ denote the location and observation of the $k$th sensor, $(\alpha_i,\mu_i)$ are uniformly drawn from the prior, $N_\mathrm{abs}={10}^5$ and the tolerance $\epsilon_\mathrm{abs}=2.5\times{10}^{-4}$. The approximate posterior densities are shown in Fig.~\ref{random_abc}.

\section{Summary}
\label{S:5}
A deep learning approach for solving forward and inverse problems involving fractional partial differential equations is presented. Using the idea of Monte Carlo quadrature and physics informed neural networks, we propose a MC-PINNs method that can flexibly compute the unbiased estimation of the FPDEs-constraint in the loss function during the optimization process of the DNNs-parameters. This approach substantially mitigates the issue of great growth in the number of auxiliary points especially for high dimensional problems, which was used in \cite{pang2019fpinns} to discretize the fractional derivative of the DNNs-output. We have demonstrated the performance of the proposed MC-PINNs for high dimensional integral fractional Laplacian, parametric identification time-space fractional differential equations and fractional diffusion equation with random inputs. Future applications and explorations include extending this approach to more general nonlocal problems and peridynamic models.
%

\FloatBarrier

\bibliography{reference_MC-Vpinn}

\appendix

\end{document}